\pdfoutput=1

\documentclass[11pt]{article}

\usepackage[final]{coling}

\usepackage{times}
\usepackage{latexsym}

\usepackage[T1]{fontenc}

\usepackage[utf8]{inputenc}

\usepackage{microtype}

\usepackage{inconsolata}

\usepackage{graphicx}
\usepackage{booktabs}
\usepackage{mwe}
\usepackage{amsmath,amssymb,amsfonts}%
\usepackage{mathrsfs}%
\usepackage[title]{appendix}%
\usepackage{xcolor}%
\usepackage{textcomp}%
\usepackage{manyfoot}%
\usepackage{algorithm}%
\usepackage{algorithmicx}%
\usepackage{algpseudocode}%
\usepackage{listings}%
\usepackage{inconsolata}

\usepackage{times}
\usepackage{latexsym}
\usepackage{paralist}
\usepackage[utf8]{inputenc}
\usepackage{microtype}
\usepackage{multirow}
\usepackage{caption}
\usepackage{subcaption}
\usepackage[rightcaption]{sidecap}
\sidecaptionvpos{table}{c}
\sidecaptionvpos{figure}{c}
\usepackage{wrapfig,lipsum}
\usepackage{hyperref}
\usepackage{amssymb}
\usepackage{amsmath}
\usepackage{amsthm}
\usepackage{enumitem}
\usepackage{pifont}
\usepackage{mdframed}
\usepackage{array}
\newcommand{\sysname}[1]{\textsc{RA-MtR}}

\definecolor{ForestGreen}{RGB}{34,139,34}
\newcommand{\cmark}{\textcolor{ForestGreen}{\ding{52}}}%
\newcommand{\xmark}{\textcolor{red}{\ding{55}}}%

\newcommand{\sa}[1]{\textcolor{black}}
\title{\sysname{}: A Retrieval Augmented Multi-Task Reader based Approach for Inspirational Quote Extraction from Long Documents }

\author{Sayantan Adak \and Animesh Mukherjee\\
  Indian Institute of Technology Kharagpur, West Bengal -- 721302\\
   \small{\href{mailto:sayantanadak.skni@kgpian.iitkgp.ac.in}{{sayantanadak.skni@kgpian.iitkgp.ac.in}}, \href{mailto:animeshm@cse.iitkgp.ac.in}{{animeshm@cse.iitkgp.ac.in}}
 }
}

\begin{document}
\maketitle
\begin{abstract}
Inspirational quotes from famous individuals are often used to convey thoughts in news articles, essays, and everyday conversations. In this paper, we propose a novel context-based quote extraction system that aims to extract the most relevant quote from a long text. We formulate this quote extraction as an open domain question answering problem first by employing a vector-store based retriever and then applying a multi-task reader. We curate three context-based quote extraction datasets and introduce a novel multi-task framework \sysname{} that improves the state-of-the-art performance, achieving a maximum improvement of 5.08\% in BoW F1-score. \footnote{Code and Data are available at \url{https://github.com/sayantan11995/Context_based_Quote_Extraction}}
\end{abstract}

\section{Introduction}

\label{sec1}

\begin{figure}[ht]
    \centering
    \includegraphics[width=0.9\linewidth]{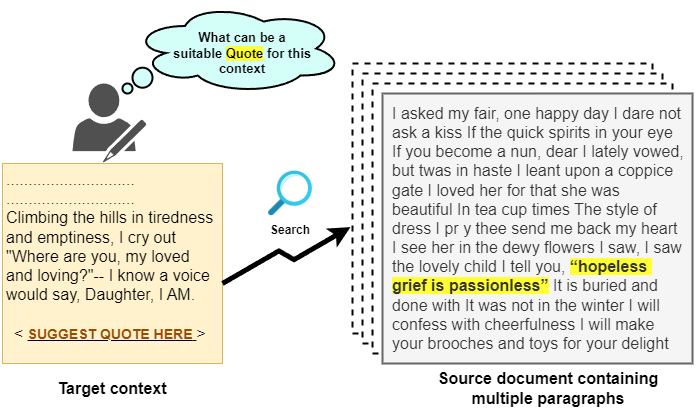}
    \caption{\footnotesize Example use-case of context-aware quote extraction from source document while composing an article. The highlighted portion from the source document can be a suitable quote for the target context in the left.}
    \label{fig:introduction}
    \vspace{-0.5em}
\end{figure}

Inspirational quotes from famous individuals are powerful tools that convey wisdom and insight in a concise and often figurative manner. They provide a secondary voice that reinforces the author's thoughts and beliefs  \cite{liu-etal-2019-neural-based}. Context-aware quote extraction (also known as quote recommendation) is crucial in writing news articles, blogs, and summaries, as it helps to strengthen the expressed ideas. This process involves identifying phrases or sentences within a paragraph that are quotable and determining their relevance and quotability in a given context. Since ``context'' can be highly subjective, finding the most relevant quotes can be challenging due to the linguistic nuances involved. Figure~\ref{fig:introduction} demonstrates a recommendation for a quotable phrase from a source paragraph, based on one context from the example of our dataset.
It turns out that authors have to spend far too much time deciding what-to-quote from many source texts analyzing their context.  Accordingly, it is in significant demand to automate the process of extracting quotes from a text. 

To tackle the task, \citet{bendersky-smith-2012-dictionary} attempts to identify ``quotable'' phrase from books on the basis of linguistic and rhetorical properties. Unlike this, (\citet{tan2015learning}, \citet{10.1145/2983323.2983788, tan}, \citet{quoteR}), leverage ``context'' to select the most relevant quote from a list of quotes. (\citet{lee, wang_trans}) use dialogue history as the \textit{context}. The task of finding the most relevant quote itself remains challenging. Moreover, our task poses inherent difficulty, as we not only attempt to find the most relevant quote for a given context, but also extract the quote from a full source document (containing several hundreds of paragraphs). To the best of our knowledge, only \citet{maclaughlin2021context} attempts to extract context-aware quotes from text documents (US presidential speech transcripts). However, the length of the documents are considerably small (see Table~\ref{tab:dataset_stat} for details) and they only cover the political domain for quote extraction. In addition, none of the experimental dataset apart from \citet{quoteR} is publicly available.

In this research, we focus on bridging the gap by rigorously curating three datasets for context aware quote extraction task, and presenting a novel framework that can enhance the task of extracting quotes from a much longer text. Our contributions can be summarized as follows.
\begin{compactitem}
    \item To better extract quotes based on the context, we propose a Retrieval Augmented Multi-task Reader (\sysname{}) framework that utilizes a vector-store for initial retrieval followed by \textit{Llama-3} based re-ranker, and a multi-task framework that leverages two training tasks tailored specifically for the quote extraction scenario.
    
    \item \textcolor{black}{ We curate two datasets for context-based quote extraction by adapting two existing quote recommendation resources—QuoteR, predominantly featuring literary quotes~\cite{quoteR}, and Quotus, which includes quotes from political speeches~\cite{niculae2015quotus}. Additionally, we introduce a dataset centered on quotes from Mahatma Gandhi. Together, these datasets span diverse genres, and we make them publicly available to foster further research in this domain.}
    
    \item \textcolor{black}{We conduct rigorous experiments using \sysname{} and show that our framework outperforms the best-performing baseline by a maximum of \textbf{5.08\%} in terms of BoW F1-score while considering the top-ranked paragraph as the location of the quote. 
    \item We also perform analysis with the \textit{multi-task reader} to demonstrate that our fine-tuned multi-task framework based on SpanBERT $\oplus$ SpanBERT-CRF improves the quote predictions over a series of baselines. Our multi-task framework outperforms the standard BERT-based models by a large margin in a few shot settings. In particular, we see that even with eight data points from the target domain, our model beats BERT and SpanBERT by \textbf{14\%} and \textbf{11\%}  in BoW F1-score respectively (see Table~\ref{tab:few-shot}.)} 
    
\end{compactitem}

\section{Related work}
\noindent\textbf{Quotability detection}:
\textcolor{black}{\citet{bendersky-smith-2012-dictionary} developed a quotable phrase extraction process that includes a supervised quotable phrase detection using lexical and syntactic features. \citet{wang_trans} introduced a transformation matrix that directly maps the query representations to quotation representations. \citet{maclaughlin-smith-2021-content} utilized BERT-based models for ranking the quotable paragraphs while evaluating on five different datasets. \citet{vosk} discussed challenges of retrieving news articles in the context of developing event-centric narratives.} 

\noindent\textbf{Context based quote recommendation}:
\textcolor{black}{\citet{tan2015learning} proposed a learning-to-rank framework for quote recommendation. 
\citet{10.1145/2983323.2983788} proposed a quote recommendation framework by learning the distributed meaning representations for the contexts and the quotes using LSTM. \citet{lee} built a quote recommender system to predict quotes based on Twitter dialogues as context. 
\citet{quoteR} built a large and the first publicly available dataset for quote recommendation. \citet{maclaughlin2021context} attempted to simultaneously rank the most quotable paragraphs and predict the most quotable spans from source transcripts modeling quote recommendation as an openQA problem.}\\ 
\if{0}\noindent\textbf{RAG and multi-task learning for openQA}: \textcolor{black}{Recently, retrieval augmented generation (RAG) based approach receives huge attention for different knowledge intensive NLP tasks. \citet{10.1145/3626772.3657834} thoroughly examines the importance of retrieval phase in a RAG based system. \citet{kim2024reragimprovingopendomainqa,nian2024wragweaklysuperviseddense} employ RAG based method for openQA tasks.  Multi-task learning has been widely used in the context of reading comprehension tasks, where multiple complementary tasks are simultaneously learnt to improve the main task's performance. Few lines of work \cite{10.1145/3269206.3271702, nandy-etal-2021-question-answering, wang-etal-2021-retrieval} jointly trained retriever and reader using embedding models to predict final answer given a question. \citet{kongyoung-etal-2022-monoqa} used encoder-decoder based model to extract answer spans. However, multi-task learning has not been explored in the domain of quotes extraction.}\\ 
\fi
\noindent\textbf{The present work}: We extend the work of \citet{maclaughlin2021context}, by proposing a novel retriever augmented multi-task reader based quote extraction. The framework employs a \textit{vector-store} based paragraph retriever followed by a decoder-only transformer based re-ranker and a novel multi-task based reader containing a sequence tagging module for identifying quotable phrases along with context aware span prediction. We curate three datasets of different genres and evaluate our approach. Our method outperforms all the previous baselines and generalizes better in a cross-domain few-shot setting.

\section{Approach}
We formalize the problem as an open-QA task, similar to the one described in \citet{maclaughlin2021context}. Given a target context ($T_C$), and a source document ($S_D$) which consists of several paragraphs ($ = \{P_1, P_2, .., P_n\}$), we require to first identify the most relevant list of paragraphs depending upon $T_C$, and then identify the most quotable phrase from the selected paragraphs. We propose the overall quote extraction approach consisting of a \textit{retriever} (detailed in section \ref{method:retriever}) to select the most relevant paragraph followed by a \textit{multi-task reader} (detailed in section \ref{sec:reader}) to extract a quote.

\section{Dataset curation}
In this section we present the details of the datasets first by discussing the quotes that we consider, followed by construction of \textit{source paragraph} and \textit{target context} for our experiments.
\subsection{Training data}
\noindent\textbf{QuoteR}: We primarily consider the English subset of the QuoteR dataset proposed by \citet{quoteR}, known to be the largest publicly available dataset for the quote recommendation task. The authors of the corresponding paper collected several quotes from the popular \textit{WikiQuote}\footnote{\url{https://en.wikiquote.org/}} project and search for the occurrences of these quotes in the Project Gutenberg corpus\footnote{\url{https://www.gutenberg.org/}}, the BookCorpus \cite{zhu2015aligning}, and the OpenWebText corpus \cite{Gokaslan2019OpenWeb}
respectively, and considered the preceding and the following 40 words of a particular quote to its left and right context respectively. After preprocessing, the authors provided a total of \textbf{6108} unique quotes and around \textbf{127k} contexts for those quotes.  
\subsection{Test data}
\noindent\textbf{Gandhi quotes}: Websites such as mkgandhi\footnote{\url{https://mkgandhi.org/}} has made the Collected Works of Mahatma Gandhi (CWMG) publicly available, which is a huge text corpus consisting of 100 volumes. We collect around a total of 800 Mahatma Gandhi quotes in English from Goodreads\footnote{\url{https://www.goodreads.com/author/quotes/5810891.Mahatma\_Gandhi?page=35}} and the \textit{mkgandhi} portal.\\
\noindent\textbf{Quotus data}: The authors in~\citet{maclaughlin2021context} utilizes the \textit{Quotus} \cite{niculae2015quotus} dataset for their experiments. The dataset consists of two sets of texts -- transcripts of US Presidential speeches  and press conferences from 2009-2014 (\textit{the source document}), and news articles that report on the speeches (\textit{the target document}). The authors crawled the articles and transcripts from the provided links in the Quotus data, and preprocessed them to gather a significant amount of quote, contexts, and paragraphs. However, they did not make their dataset publicly available. We ourselves re-scraped the links from the source Quotus data.\\
\noindent\textbf{Curating source paragraph and target context}: \textcolor{black}{From these three dataset (i.e., \textit{QuoteR}, \textit{Gandhi}, and \textit{Quotus}) we get a list of quotes. However, to evaluate our \textit{retriever} and \textit{reader}, we require to curate the source paragraph and the target context for each of the quotes. We leverage the \textit{Project Gutenberg} corpus to construct \underline{4,889} quote-context-paragraph triples (containing \underline{1,708} unique quotes) for QuoteR. We use \textit{Gadhipedia}\footnote{\url{https://www.gandhipedia150.in}} search engine to curate \underline{737} triples for the Gandhi quotes. For the Quotus, we utilize the \textit{Quoting POTUS}\footnote{\url{http://snap.stanford.edu/quotus/vis/}} website containing the news articles and align the quotes to source transcript for constructing \underline{2,698} triples. The detailed steps and algorithms are provided in Appendix~\ref{appendix:data_preprocessing}.} 
\subsection{Dataset statistics}
\begin{figure*}[th]
    \centering
    \begin{subfigure}[b]{0.32\textwidth}
        \label{Fig:quoter_word_cloud}
        \centering
        \includegraphics[width=\textwidth]{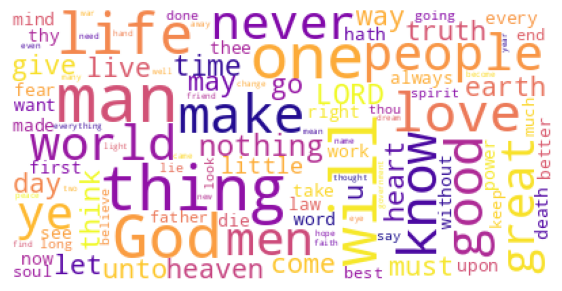}
        \caption{\footnotesize QuoteR}
    \end{subfigure}
    \begin{subfigure}[b]{0.32\textwidth}
        \label{Fig:gandhi_word_cloud}
        \centering
        \includegraphics[width=\textwidth]{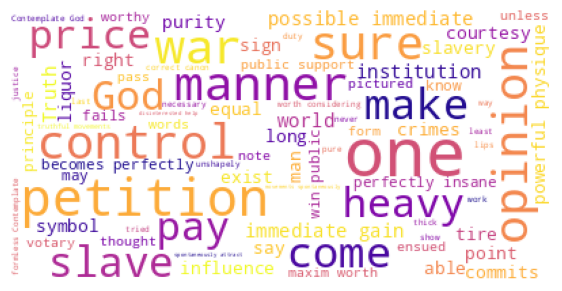}
        \caption{\footnotesize Gandhi}
    \end{subfigure}
    \begin{subfigure}[b]{0.32\textwidth}
        \label{Fig:quotus_word_cloud}
        \centering
        \includegraphics[width=\textwidth]{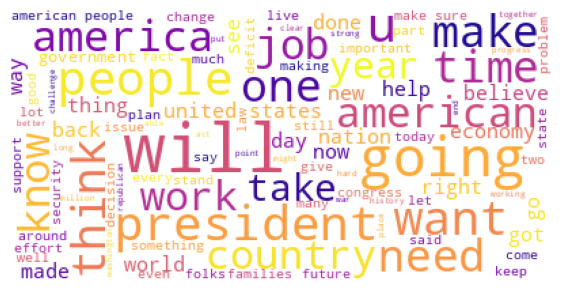}
        \caption{\footnotesize Quotus}
    \end{subfigure}
    \caption{\footnotesize Most prominent words present in the quotes across the three datasets.}
    \label{fig:word_cloud_quote}
\end{figure*}

Thus, overall we consider \textit{three} datasets each from a different genre - (i) QuoteR - where most of  the quotes are from novels, 2) Gandhi - solely based on the quotes of Mahatma Gandhi, and 3) Quotus - quotes from the political speech. The basic statistics of these three datasets are noted in Table~\ref{tab:dataset_stat}. Figure~\ref{fig:word_cloud_quote} demonstrates the most prominent words present in the three datasets. The quotes in the QuoteR and Gandhi datasets contain positive words like ``God'', ``good'', ``love'', ``truth'', ``petition'' etc. The Quotus dataset on the other hand contains quotes having words ``america'', ``president'' etc. We also compare our dataset with the dataset used in other similar works (see Table~\ref{tab:dataset_comparison}). We present the only dataset containing quote, context and source paragraph. These datasets will be made publicly available upon acceptance.

\begin{table}[h]
\scriptsize
\centering
\resizebox{\columnwidth}{!}{
\begin{tabular}
{l|p{0.10\columnwidth}|p{0.15\linewidth}|p{0.10\linewidth}|p{0.10\linewidth}|p{0.12\linewidth}|p{0.12\linewidth}}\toprule
\textbf{Dataset} &\textbf{\# of unique quotes} &\textbf{\# of quote-context-paragraph triples} &\textbf{Avg. \# of tokens / quote} &\textbf{Median \# of tokens / quote} &\textbf{Avg \# of para / Src} &\textbf{Avg \# of tokens / Src} \\\hline
\textbf{QuoteR}  & 1708 & 4889 & 13.51 & 11 &551 & 98783.17 \\
\textbf{Gandhi} & 737 &737 &20.42 &19 &19.54 &3812.47 \\
\textbf{Quotus} &2698 &2698 &20.46 &16 &86.78 &4631.55 \\
\bottomrule
\end{tabular}
}
\caption{\footnotesize Statistics for the three datasets. For QuoteR we report the instances that we could find in the Gutenberg corpus.}\label{tab:dataset_stat}
\end{table}

\begin{table}[!htp]\centering
\scriptsize
\resizebox{\columnwidth}{!}{
\begin{tabular}{l|c|c|c|c}\toprule
\textbf{Dataset} &\textbf{Context based} &\textbf{Context type} &\textbf{Source paragraph}  &\textbf{Public}\\\midrule
\citet{bendersky-smith-2012-dictionary}  &\xmark &writings &\xmark  &\xmark\\
\citet{tan2015learning} &\cmark &writings &\xmark &\xmark\\
\citet{wang_trans}  &\cmark &dialogue &\xmark  &\cmark\\
\citet{quoteR}  &\cmark &writings &\xmark  &\cmark\\
\citet{maclaughlin2021context} &\cmark &writings &\cmark &\xmark\\
Our dataset &\cmark &writings &\cmark &\cmark\\
\bottomrule
\end{tabular}}
\caption{\footnotesize{Dataset comparison for other related tasks with ours.}}
\label{tab:dataset_comparison}
\end{table}

\section{Methodology}
\begin{figure*}[ht]
    \centering
    \includegraphics[width=0.9\linewidth]{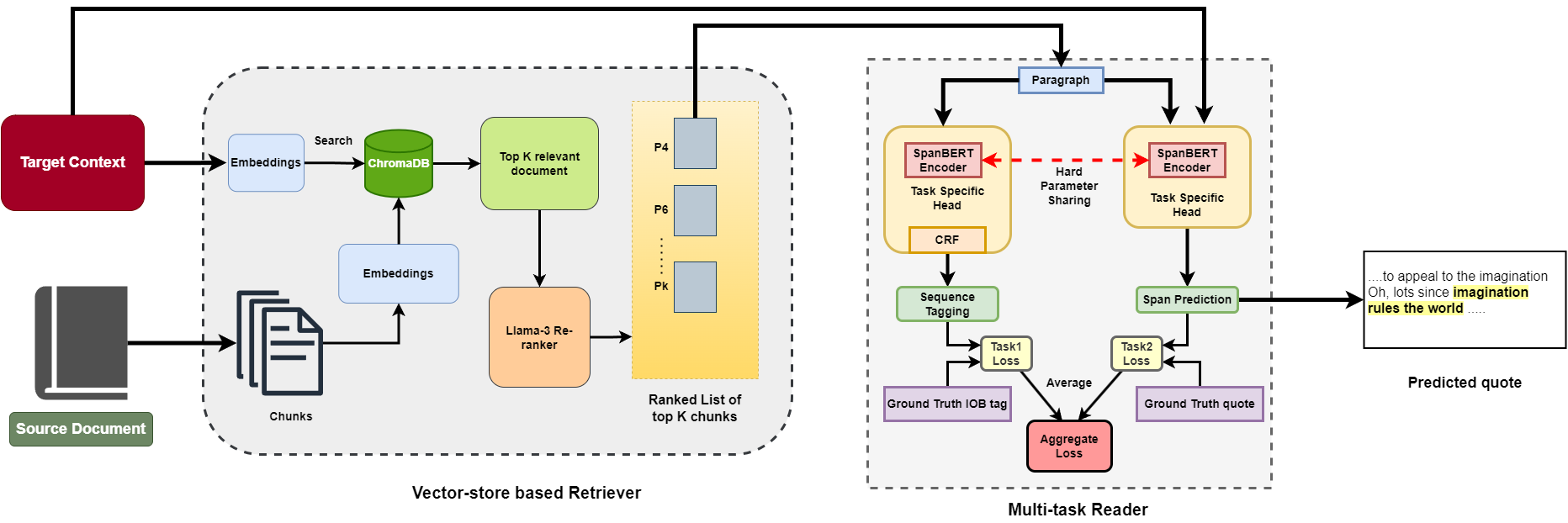}
    \caption{\footnotesize The \sysname{} architecture.}
    \label{fig:RA_MTR_Full}
\end{figure*}

In this section, we describe the details of our methodology for quote extraction. We propose the overall quote extraction approach as an open-QA framework, which normally consists of a \textit{retriever} and a \textit{reader}. The retriever is essential for selecting the top paragraphs relevant to the context from the whole document. We employ a novel multi-task learning framework in the reader, which extracts the most quotable spans from the selected paragraphs and is discussed in detail below. The overall retriever-reader architecture \sysname{} is illustrated in the left part of  Figure~\ref{fig:RA_MTR_Full}.


\subsection{\sysname{}: Retriever module}
\label{method:retriever}
Inspired by the RAG \cite{NEURIPS2020_6b493230} architecture, we employ a vector-store based retriever to initially retrieve top-$k$ ($k=20$)\footnote{Increasing \textit{k} did not change the performance too much.} paragraph based on the given context. We utilize \textit{langchain} API\footnote{\url{https://python.langchain.com/docs/modules/data\_connection/}}, to split the source document into several chunks (we choose chunk-size of 1200\footnote{We find maximum length of context + paragraph is 1005.} characters and chunk-window of 100), followed by encoding of each chunk using \textit{sentence-transformers}, and finally store the embeddings into a vector-store for efficient searching. We use ChromaDB\footnote{\url{https://pypi.org/project/chromadb/}} for storing the embeddings of the chunks. In parallel, the query context is also embedded using \textit{sentence-transformers}. To extract the relevant inspirational quote from the source document we perform a similarity search by comparing the query context embedding and the embeddings in the vector-store to retrieve top-$k$ chunks. The retrieved chunks are then passed to the more powerful
sequence-to-sequence re-ranking module for further processing.\\
\noindent\textit{Fine-tuning paragraph re-ranking module}: \textcolor{black}{After retrieving initial sets of candidate paragraphs, many past literature leveraged deep neural network based paragraph re-ranking modules to get a final ranked list of paragraphs. Works such as  \citet{dai2019deeper, yilmaz2019applying, nogueira2019multi} exploited BERT for paragraph/document retrieval tasks. \citet{nogueira-etal-2020-document} used a T5-ased encoder-decoder architecture for document ranking. We apply a more sophisticated decoder-only transformer based model \textit{Llama-3}\footnote{We apply the chat model from huggingface \url{meta-llama/Meta-Llama-3-8B-Instruct}} to re-rank the paragraph. Similar to \citet{nogueira-etal-2020-document} we formulate the problem as a binary classification task, and the input prompt is:
\begin{mdframed}[]
\scriptsize
Context:\quad \{c\}\\
Document:\quad \{d\}\\
Is the document relevant to the context? Answer yes/no: 
\end{mdframed}
\noindent where $c$ and $d$ are the context and paragraph texts, respectively. The model is fine-tuned to produce the words \texttt{yes} or \texttt{no} depending on whether the document is relevant or not to the query. That is, \texttt{yes} and \texttt{no} are the `target words' (i.e., ground truth predictions in the sequence-to-sequence transformation). To generate training and test examples for the models, we iterate over each context
and create (context, source paragraph, label) example triples for each paragraph in the corresponding source document. The label is \texttt{yes} if the author actually quoted from the paragraph (positive triple) and \texttt{no} (negative triple) otherwise. 
At inference time, to compute probabilities for each query–document pair (in a re-ranking setting), we retrieve the unprocessed next-token probabilities for the tokens \texttt{yes} and \texttt{no}. From these, we calculate the  $yes-score$ as follows. 
\begin{equation}
yes-score_{(c, d_i)} = \frac{p(yes|P)}{p(yes|P) + p(no|P)}
\end{equation}
where, $c$ is the context, $d_i$ is the $i^\textrm{th}$ document and $P$ is the prompt. Similarly, as baseline, we also fine-tune encoder-decoder based models (T5, FLAN-T5) for the re-ranking task using similar approach.}


\noindent\textit{Sampling hard negatives}: Out of all the negative triples obtained we select $n$ hard samples for training to make the model more robust. We explore two different hard negative sampling methods - a) select the paragraphs that are closest next to the positive paragraph, and b) select the top ranked paragraphs (other than the positive one) returned by BM25 retriever model. However, we observe that both choices produce similar results (see results section for details). 

\subsection{\sysname{}: Multi-task reader module}\label{sec:reader}
\begin{figure}[ht]
    \centering
    \fbox{\includegraphics[width=0.75\columnwidth]{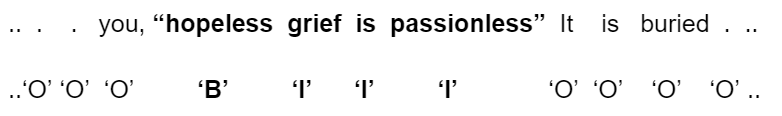}}
    \caption{\footnotesize Example of BIO tagging.}
    \label{fig:BIO_tagging}
\end{figure}

\noindent\textbf{Motivation for multi-task training}: \textcolor{black}{Unlike normal spans of text, quotes have certain inherent special properties or some figurative language that make them unique \cite{bendersky-smith-2012-dictionary}. We believe that identifying such special occurrences of phrases is essential for quote prediction from paragraphs. We cast this as a sequence tagging, i.e., marking only the portions of a text that can be recounted as quotable. We attempt to optimize two tasks in parallel - quotable sequence tagging (using SpanBERT-CRF) and context aware span prediction (using SpanBERT). In a paragraph, there can be multiple spans of text which will be relevant to the context. However, not every relevant span is quotable. The span prediction module is essentially a variant of a question-answering module, which might not be good enough to identify quotability of the answer. Also, many of the quotes are only subparts of a sentence (e.g., ``He travels fastest who travels alone,...'') while few of the quotes consist of more than one sentence (e.g., ``In this world there are only two tragedies. One is not getting what one wants, and the other is getting it.''). To mitigate this gap, we use a specific sequence identification module (SpanBERT-CRF) to find quotable sequences.}\\
\noindent\textbf{Span prediction from paragraph}: We train the span prediction model using context-quote-paragraph triple as the training example. Similar to \citet{maclaughlin2021context}, we utilize the span-level BERT architecture, which receives the packed sequence of the context and paragraph as input. By utilizing the final hidden
vector \(T_i \in \mathbb{R}^{h} \)
as the representation for each wordpiece
in a given paragraph, we follow the standard approach of casting span prediction as two classification tasks, i.e., separately predicting the start and end of the span \cite{devlin-etal-2019-bert}. To this purpose, we introduce a start vector, \( S \in \mathbb{R}^{h} \), and an end vector, \(E \in \mathbb{R}^{h} \). The probability of a word $w$ being the start of the quoted span is the dot product \(S \cdot T_w\) followed by a softmax over all wordpieces in the example. We follow the same approach for calculating the probability of word $w$ being the end of the span using $E$. The loss is calculated as the average NLL (Negative  log-likelihoods) of the correct start and end positions, i.e., the tokens in the paragraph the author actually quoted. Following \citet{devlin-etal-2019-bert}, at prediction time, the score of the span from position $i$ to $j$ is \(S \cdot T_i+E \cdot T_j\). We consider a span
valid if $j > i$ and $i$ and $j$ occur in the paragraph portion of
the input. We retain a mapping from wordpieces to original tokens for prediction.\\
\noindent\textbf{Quotability detection as sequence tagging}: \citet{scheible-etal-2016-model, pareti-etal-2013-automatically} framed the quotation detection task as sequence tagging. \citet{portelli2021improving} used sequence labeling for the adverse drug events (ADE) detection from a given text. Along similar lines, we employ SpanBERT neural model combined with Conditional Random Field (CRF) to identify quotable phrases or words.  
Each example from the dataset is accompanied by a paragraph, the start and end character positions of the quote in that paragraph. Using this information, we first convert this into the commonly used IOB (Inside, Outside, Beginning) schema using \textit{Spacy}. 
Consider the example in Figure~\ref{fig:introduction}, every word except the bold portion (i.e., the quote) should be marked as `O'. The word `hopeless' in the quote should be labeled as `B' and the rest of the quote should be labeled as 'I'. The BIO tagging is illustrated in Figure~\ref{fig:BIO_tagging}.
Since BERT-based models generally employ wordpiece tokenizers to tackle the out-of-vocabulary words, which actually break such words into multiple subwords, we require to decide on a consistent IOB schema for the subwords. We set a rule for the sub-labels which are consistent with the IOB schema: words labeled as B generate a series of subwords labeled as [B, I, . . . , I], while words labeled as I (or
O) generate a series of identical I (or O) sub-labels. For example, if the word `resisted' has the label B, then its corresponding wordpieces - [`Resist', `\#\#ed'] would get labeled as [B, I].\\ 
\noindent\textbf{Multi-task training}: To take advantage of both the span prediction model and the quotable phrase identification model, we adopt a multi-task based framework where we have two independent models and they share the same transformer encoder. The span prediction model tries to match the start and end token of the quote in the paragraph, whereas the quotable phrase identification model tries to predict the `B', `I', and `O' labels for each token. During the backpropagation, we average the losses from the two models. The right part in the Figure~\ref{fig:RA_MTR_Full} demonstrates the architecture of the multi-task framework.

\begin{table}[!t]\centering
\resizebox{\columnwidth}{!}{
\begin{tabular}{
>{\raggedright\arraybackslash}p{15em}
|cc|cc|ccc}\toprule
\multirow{3}{*}{\textbf{Method}} &\multicolumn{6}{c}{\textbf{Dataset}} \\
\cline{2-7}
&\multicolumn{2}{c|}{\textbf{QuoteR (test)}} &\multicolumn{2}{c|}{\textbf{Gandhi}} &\multicolumn{2}{c}{\textbf{Quotus}} \\
&\textbf{Top-1} &\textbf{Top-3} &\textbf{Top-1} &\textbf{Top-3} &\textbf{Top-1} &\textbf{Top-3} \\\hline
DrQA \cite{chen2017reading} &7.19 &8.22 &6.20 &8.38 &4.26 &5.43 \\
ParagraphRanker \cite{lee-etal-2018-ranking} &16.58 &21.45 &12.17 &14.35 &11.31 &15.11 \\
BM25 + \cite{maclaughlin2021context} (Positive only settings) &31.78 &37.37 &23.70 &26.60 &32.58 &37.68 \\
Contriever + FiD &32.81 &39.22 &23.78 &28.25 &33.36 &39.43 \\
BM25 + BERT-base + MTR* &{37.20} &{46.28} &{25.17} &{27.30} &{36.15} &{41.12} \\
BM25 + T5-base + MTR* &{34.17} &{45.21} &{21.31} &{23.45} &{37.13} &{39.25} \\
BM25 + T5-large + MTR* &{39.07} &{48.29} &{28.13} &{30.67} &{39.29} &{41.11} \\
BM25 + FLAN-T5-large + MTR* &{43.12} &{51.43} &{34.46} &{42.30} &{42.26} &{48.21} \\
Vector-store based retriever + LLM reader (Llama-3-8b-instruct) &{14.81} &{18.76} &{17.53} &{23.28} &{31.38} &{39.77} \\
\textbf{\sysname{} (ours)} &\textbf{45.74} &\textbf{57.25} &\textbf{38.71} &\textbf{49.38} &\textbf{43.40} &\textbf{53.45} \\
\bottomrule
\end{tabular}
}
\caption{\footnotesize The result (F1 score) for the quote extraction using the different baseline models and our \sysname{} approach. For a fair comparison, we took the results from the positive-only settings of \citet{maclaughlin2021context}. Note that all the fine-tuned models are only trained on the QuoteR training data. *MTR: Our fine-tuned multi-task reader. Results using different LLMs are reported in Table~\ref{tab:end-to-end-inference_LLM} in Appendix~\ref{appendix:llm_results}.}\label{tab:end-to-end-inference}
\end{table}

\section{Experiments}

In this section, we discuss the experiments that we conduct and the details of the experimental setups.\\
\noindent\textbf{Fine-tuning paragraph re-ranking}: We pass the packed input of context and paragraphs to the retriever model. Out of \textbf{4889} data points in the QuoteR dataset, and we select 80\% for training, 10\% each for dev and test. We choose to fine-tune the \textit{Llama-3-8b-instruct} model for the paragraph ranking task. For model implementation details and hyper-parameters see Appendix~\ref{appendix:model_impl}.\\
\noindent\textbf{Fine-tuning reader}: We fine-tune the reader by randomly selecting 80\% QuoteR data for training, 10\% each for dev and test (see Appendix~\ref{appendix:model_impl} for implementation details). 
To test the generalizability of the model in a few-shot setting, we consider random training samples \( \in \{4, 8, 16, 32, 64\}\) from the other two datasets (i.e., Gandhi and Quotus) for further fine-tuning with a slightly lower learning rate (1\(e^{-5}\)), and test on the remaining data samples for the respective datasets.\\
\noindent\textbf{Metrics}: Since the setup for our span prediction task is identical to QA, we evaluate the span-level models using the two popular
QA metrics  – (i) exact match (EM), and (ii) macro-averaged bag-of-words (BoW) F1. EM measures if the predicted span exactly matches the positive quote, and BoW-F1 measures their average word overlap.\\
\noindent\textbf{Baselines}: Both the retriever and the reader can have many variants which serve as ideal baselines. In the retriever part we use vanilla BM25 as a first baseline. Apart from the simple BM25 retriever, we employ encoder-based (BERT), encoder-decoder based (T5, FLAN-T5) document re-ranking to improve paragraph selection.\\ For the \textit{reader} part, as primitive baselines, we consider using the first and last sentences of each paragraph as potential quotes. To further explore, we also fine-tune the BERT and the SpanBERT pretrained models on the BERT question answering architecture. We keep the same hyperparameter settings as the multi-task framework. Additionally, we also observe the ability of different medium sized open-source LLMs such as FLAN-T5-large\footnote{\url{https://huggingface.co/google/flan-t5-large}}, FLAN-T5-XL\footnote{\url{https://huggingface.co/google/flan-t5-xl}}, Bloomz-3b\footnote{\url{https://huggingface.co/bigscience/bloomz-3b}}, Falcon-7b\footnote{\url{https://huggingface.co/tiiuae/falcon-7b}}, Llama-3-8b\footnote{\url{https://huggingface.co/ meta-llama/Meta-Llama-3-8B-Instruct}} models for zero-shot context-aware quote extraction task. For the detailed methodology, refer to Appendix~\ref{appendix:baseline}.

\begin{table}[ht]
\centering
\resizebox{0.9\columnwidth}{!}{
\begin{tabular}{l|cc}\hline
\textbf{Method} &\textbf{EM} &\textbf{BoW-F1} \\\hline
First sentence &0.55 &6.31 \\
Last sentence &1.08 &5.88 \\
BERT-base &69.1\(\pm\)0.5 & 76.2\(\pm\)0.9 \\
BERT-large &71\(\pm\)0.3 & 77.9\(\pm\)0.3 \\
SpanBERT-base &71.7\(\pm\)0.6 &77.7\(\pm\)0.4 \\
SpanBERT-large &72.3\(\pm\)0.6 &79.2\(\pm\)0.4 \\
Multi-task using SpanBERT-base (Ours) &73\(\pm\)0.8 &78.2 \(\pm\)0.3 \\
Multi-task using SpanBERT-large (Ours) &\textbf{77\(\pm\)0.4} &\textbf{86.1 \(\pm\)0.2} \\
\bottomrule
\end{tabular}
}
\caption{\footnotesize Reader performance on the QuoteR dataset. We provide the positive paragraph to predict the quote span.}\label{tab:span-prediction}

\end{table}

\begin{table}[!htp]\centering
\scriptsize
\resizebox{\columnwidth}{!}{
\begin{tabular}{l|c|ccc}\hline
\multirow{2}{*}{\textbf{Test data}} &\multirow{2}{*}{\textbf{\# training samples}} &\multicolumn{3}{c}{\textbf{Methods}} \\
& &\textbf{BERT}  &\textbf{SpanBERT} &\textbf{Multi-task (Ours)} \\\hline
\multirow{5}{*}{\textbf{Gandhi}} 
&8 &27.71  &30.30 &41.32 \\
&16 &32.60  &32.65 &50.29 \\
&32 &38.12  &36.85 &62.91 \\
&64 &43.54  &44.65 &72.08 \\\hline
\multirow{5}{*}{\textbf{Quotus}} 
&8 &33.97  &36.82 &40.58 \\
&16 &37.90  &41.84 &49.33 \\
&32 &40.56  &44.80 &55.08 \\
&64 &47.86  &51.27 &59.12 \\
\bottomrule
\end{tabular}
}

\caption{\footnotesize Few-shot inference performance on the 1) Gandhi and 2) Quotus datasets. We have used the BoW F1-score as the metric for comparison here.}\label{tab:few-shot}
\end{table}
\section{Results}
\label{section:results}

\noindent\textbf{Performance of \sysname{}}: To examine the efficacy of our entire pipeline, we conduct an end-to-end prediction from our approach. In the \textit{retriever-reader} based (baseline) approach, we first provide the context and the list of paragraphs segmented from a particular book to the paragraph retrieval module. We initially get a list of 20 top-ranked paragraphs relevant to the context from the RAG model and then re-rank these using the Llama-3 model. We take the top three paragraphs further and sequentially pass them with respect to the context to our multi-task quote extraction module. This span prediction module within the multi-task framework predicts the top three quotable spans, each from one corresponding paragraph. We measure the BoW F1-score for these three predictions with respect to the ground truth quote and report the scores for 1) top-1 prediction - score when we compare the ground truth with the predicted span from the top 1 out of the three ranked paragraphs, and 2) top-3 predictions - best score when we compare the ground truth with all the three predictions. Table~\ref{tab:end-to-end-inference} demonstrates the result for the end-to-end quote prediction \sysname{}. We compare the performance of our pipeline with two commonly used baselines for open-domain question answering tasks -- DrQA \cite{chen2017reading} and ParagraphRanker \cite{lee-etal-2018-ranking}. In addition, we employ Contriever \cite{izacard2021unsupervised} for paragraph retrieval and fine-tune a Fusion in Decoder (FiD) \cite{izacard2021leveraging} model as the reader. We also compare \sysname{} against \citet{maclaughlin2021context}, which focuses on context-based quote extraction. Lastly, we compare our model with present day LLM variants. \sysname{} by far outperforms all the baselines.   

\noindent\textbf{Multi-task reader performance}: We show the results for span prediction using various methods in Table~\ref{tab:span-prediction} for the QuoteR dataset. The results in the first two rows are from two very primitive baselines. Scores in the next two rows are from only the span selection models, which \cite{maclaughlin2021context} has considered. We can clearly see that our multi-task based approach outperforms the other methods. The improvements are significant with $p<0.05$ as per the Mann-Whitney U test \cite{mann1947test}
 and experiment with our three datasets. In Table~\ref{tab:few-shot}, we demonstrate the few-shot performances on the Gandhi and the Quotus data for the quote prediction task. We can observe that, in the few-shot settings, the multi-task framework performs much better than the simple span prediction models that are normally used for the QA tasks. In fact, with only 8 data samples from the target domain, our model beats BERT and SpanBERT by \textbf{14\%} and \textbf{11\%} for the Gandhi data, and by \textbf{7\%} and \textbf{4\%} for the Quotus data respectively. We can infer from these results that the addition of the quotable phrase identification task actually helps the model learn the linguistic properties of the quotes much better than the simple span prediction model. Further, the multi-task framework generalizes particularly well in the cross-domain setting even with the training and test paragraphs coming from different genres.

\begin{table}[!htp]\centering
\resizebox{0.85\columnwidth}{!}{
\begin{tabular}{l|cc}\hline
\textbf{Method} &\textbf{EM} &\textbf{BoW-F1} \\\hline
BERT span prediction &19.20\(\pm\)0.30 & 30.90\(\pm\)0.80 \\
SpanBERT span prediction &18.30\(\pm\)0.50 &29.70\(\pm\)0.40 \\
Multi-task (Ours) &22.00\(\pm\)0.80 &38.20\(\pm\)0.70 \\
\bottomrule
\end{tabular}
}
\caption{\footnotesize Results for the quote extraction in absence of the context (for QuoteR dataset).}\label{tab:span-prediction-without-context}
\vspace{-0.5em}
\end{table}

\noindent\textbf{Performance of the sequence tagger}: We analyze the output generated by the sequence tagger head from our multi-task framework. Note that this was an auxiliary task to improve the main task of span prediction. The sequence tagger head typically predicts `B', `I', or `O' tags for each token, and the prediction is independent of the context. We apply the model to instances in the QuoteR test dataset and extract the predicted labels from the sequence tagger head. We find that the model correctly predicts the BIO labels for 48.1\% of the instances. In 20.7\% of the cases, the model predicts multiple BIO labels within a single paragraph, indicating that one paragraph can contain multiple instances of quotable phrases.\\
\noindent\textbf{Context (in)dependence}: We conduct an ablation experiment to observe the impact of context for the quote prediction in the multi-task setting. We remove the whole context from the inputs in the test set for the quote prediction models while experimenting with the QuoteR dataset\footnote{The results from the other datasets show similar trends and hence are not shown.}. Table~\ref{tab:span-prediction-without-context} clearly shows that the baseline models' performances are drastically reduced, whereas our multi-task framework outperforms the two baselines. As the sequence tagging task is independent of the context we observe that the multi-task framework performs better in the absence of context while the two other models that are highly context-sensitive. We can infer that the linguistic boundary identification for the quotes in terms of the BIO markers enhances the performance and makes it robust to the absence of context. This is one of the prime strengths of the multi-task framework.

\section{Analysis of retriever}
\label{appendix:retriever_analysis}
\subsection{Quantitative analysis of retriever}
\textcolor{black}{We attempt to measure how much re-ranking is effective in ranking the paragraphs based on the context. We use the curated set for paragraph ranking (the QuoteR test set, and other two datasets including the source) to perform the evaluation. As baselines we use BM25 based, sentence-bert similarity based and contriever based retriever. Further we use Flan-T5 and Llama-3 based re-ranking to evaluate the importance of re-ranking. From the retriever we first take top-20 (which we use during the main experiment) retrieved paragraphs from the source and then use re-ranking to measure Precision@k (k $\in \{1,3\}$). The result is reported in Table~\ref{tab:quantitative_analysis_retriever}. We can observe that the re-ranking drastically improves the retriever performance.}

\begin{table}[!htp]\centering
\resizebox{\columnwidth}{!}{
\begin{tabular}{l|rr|rr|rrr}\toprule
\multirow{3}{*}{\textbf{Method}} &\multicolumn{6}{c}{\textbf{Dataset}} \\\cmidrule{2-7}
&\multicolumn{2}{c|}{\textbf{QuoteR}} &\multicolumn{2}{c|}{\textbf{Gandhi}} &\multicolumn{2}{c}{\textbf{Quotus}} \\\cmidrule{2-7}
&P@1 &P@3 &P@1 &P@3 &P@1 &P@3 \\\midrule
BM25 &0.15 &0.18 &0.09 &0.16 &0.13 &0.19 \\
sentence-bert similarity &0.24 &0.29 &0.14 &0.17 &0.23 &0.27 \\
contriever-based &0.27 &0.33 &0.15 &0.19 &0.25 &0.3 \\
sentence-bert + Flan-T5 re-ranking &0.43 &0.52 &0.38 &0.42 &0.44 &0.53 \\
sentence-bert + Llama-3 re-ranking &\textbf{0.48} &\textbf{0.57} &\textbf{0.41} &\textbf{0.44} &\textbf{0.47} &\textbf{0.54} \\
\bottomrule
\end{tabular}
}
\caption{\footnotesize Analysis of retriever performance along with different reranker.}\label{tab:quantitative_analysis_retriever}
\end{table}

\subsection{Qualitative analysis of different re-ranking methods}
\begin{figure}[h]
    \centering
\includegraphics[width=0.9\columnwidth]{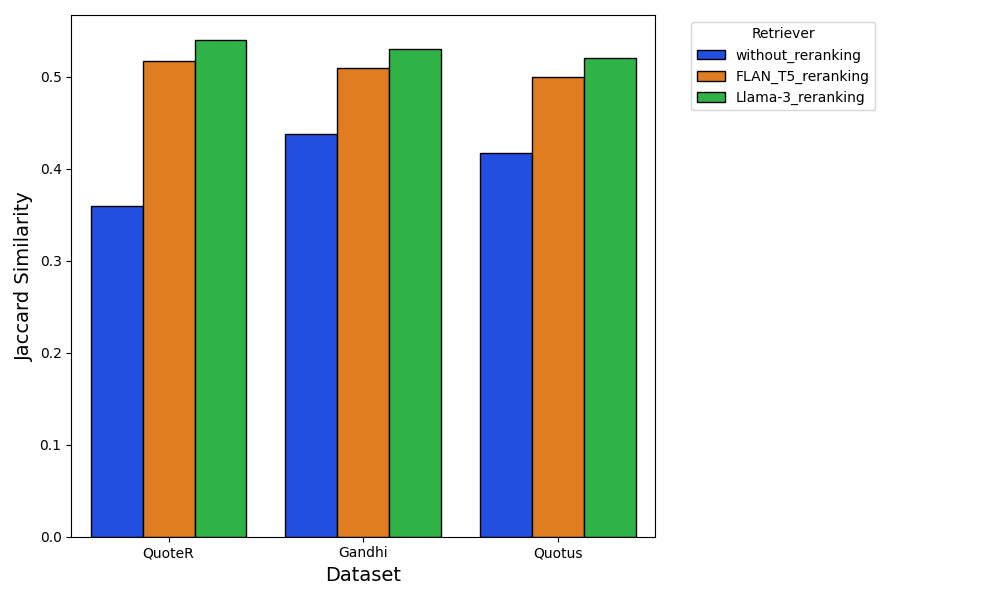}
    \caption{\footnotesize Average \textit{Jaccard} similarity between top predicted chunk and positive paragraph for a specific context.}
    \label{fig:jaccard_similarity}
\end{figure}

In this section, we attempt to analyse how well our vector-store based retriever performed.
As we cannot directly compare the retrieved chunks with the positive paragraph in our dataset (due to variable word length), we measure the average \textit{Jaccard} similarity between the top predicted chunk with the positive paragraph in our dataset for a specific context. We present the results in Figure~\ref{fig:jaccard_similarity}.
We observe that, using Llama-3 based re-ranking, the similarity significantly improved for all the three datasets. 

\section{Analysis of the multi-task reader output}
\label{sec:analysis_reader}
\begin{figure}[ht]
    \centering
    \fbox{\includegraphics[width=0.9\columnwidth]{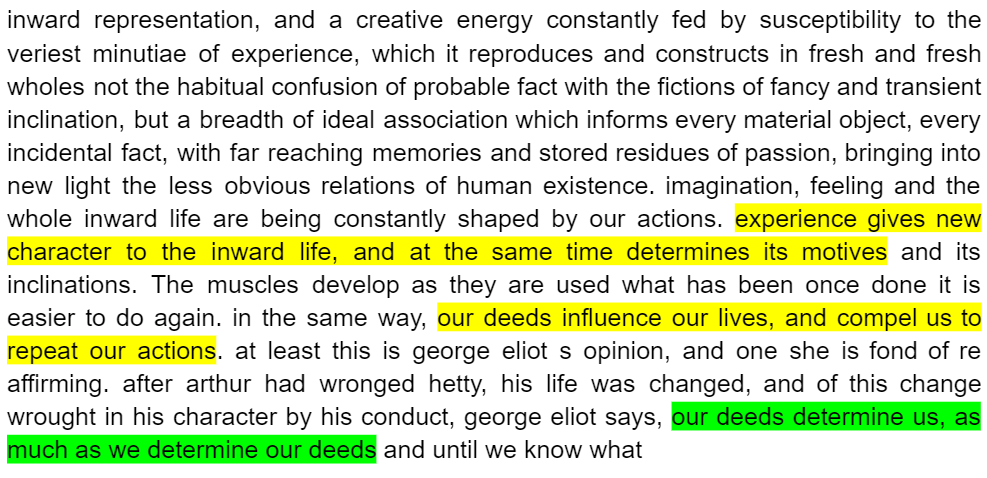}}
    \caption{\footnotesize Example of an instance where the sequence tagger wrongly predicts the BIO labels. True and predicted quotes are highlighted in green and yellow respectively.}
    \label{fig:sequence_classification_error}
\end{figure}

\noindent\textit{Analysis of top predicted quotes}: Since there may be multiple quotes in a given paragraph for a given context, we also look at the top five predicted spans from our multi-task framework for each of the paragraphs in the test set. We manually annotate the relevance of the predicted spans for the top five predictions. We had two annotators, and each of them was provided with a set of context and the top five predicted spans. They were required to mark 1 if the predicted span is semantically coherent with the context, and 0 otherwise. In the case of ambiguity, a third annotator was involved to adjudicate. We obtain an inter-annotator agreement of Cohen's \(\kappa=0.64\). We take the final relevancy (i.e., 0 or 1) based on majority vote. We achieve a high MAP@5 score of 0.78, indicating that our multi-task framework retrieved $\sim 3.9$ (on average) meaningful recommendations among the top five recommendations.\\ 
\noindent\textit{Error in the sequence tagger}: We review some instances where the model failed to predict the correct BIO labels. A specific example is depicted in Figure~\ref{fig:sequence_classification_error}, where the true quote is highlighted in green, while the predicted quotes are highlighted in yellow. Although the true and the predicted quotes come from different portions of the paragraph, they both are highly quotable as per human experts. We observe many such cases of (pseudo) errors that manifest due to the absence of valid additional ground truth quotes.\\
\noindent\textit{Error in the multi-task reader}: We examine the predicted quotes, which do not entirely match with the ground truth quotes. We observe that in 72\% the model predicts a sub-phrase of the original quote. For instance, while the actual quote is `Our Father, which art in heaven, hallowed be thy name', the corresponding predicted quote is `which art in heaven, Hallowed be thy Name'.  In a few cases, the model over-predicts, i.e., predicts a span containing the true quote and some phrases surrounding it. For example, the actual quote `Money begets money', is predicted as `Money begets money and its offspring'.

\section{Additional results}
\begin{figure}[h]
    \centering
    \begin{subfigure}[b]{0.48\columnwidth}
        \label{Fig:cosine_similarity_last_hidden_states}
        \centering
        \includegraphics[width=0.85\textwidth]{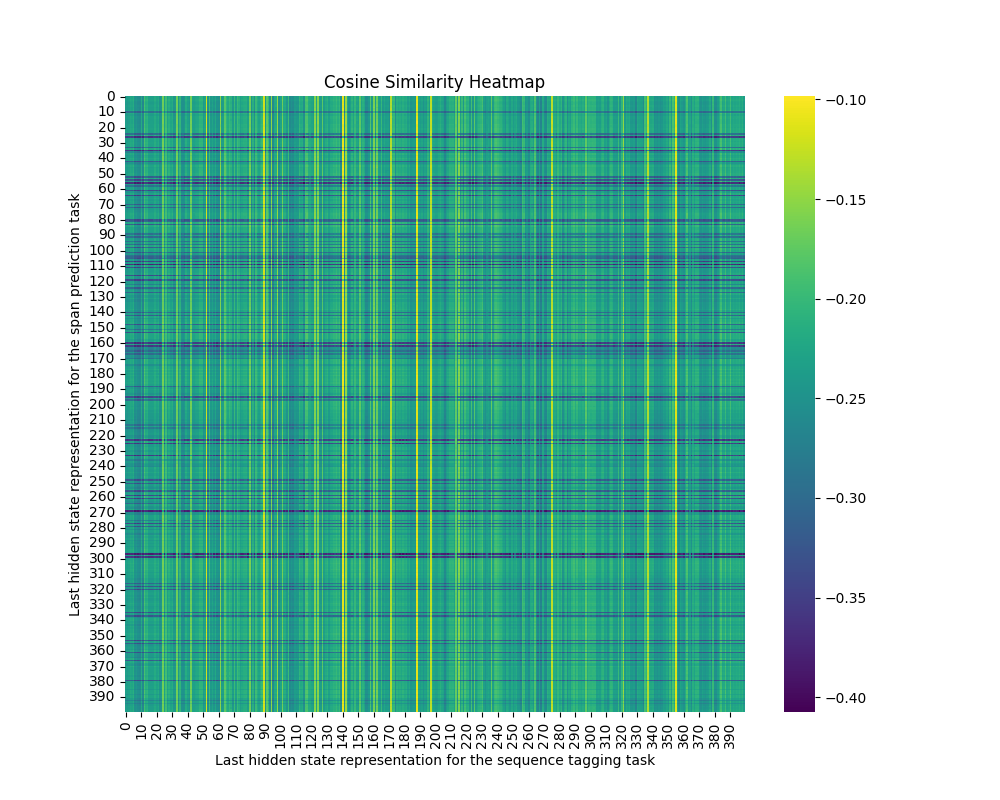}
        \caption{\footnotesize Similarity distribution}
    \end{subfigure}
    \begin{subfigure}[b]{0.48\columnwidth}
        \label{Fig:t-sne_plot_last_hidden_state}
        \centering
        \includegraphics[width=\textwidth]{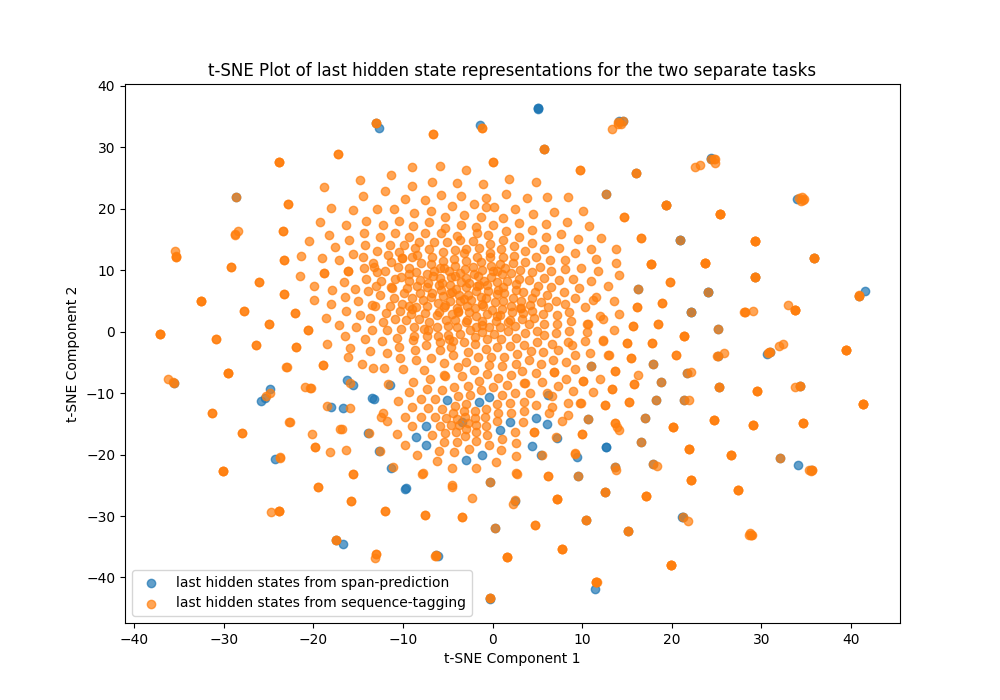}
        \caption{\footnotesize T-sne plot }
    \end{subfigure}
    \caption{\footnotesize Vector space analysis of the two tasks presented in multi-task reader.}
    \label{fig:vector_space_analysis}
\end{figure}

\noindent\textit{Ablation experiments}:\textcolor{black}{We conduct ablation experiments for the effectiveness of different components in the \sysname{} framework. We report the results for the QuoteR test set in Table~\ref{tab:ablation_analysis} (see Appendix~\ref{appendix:retriever_analysis} for more results).}\\
\begin{table}[!htp]\centering
\resizebox{0.9\columnwidth}{!}{
\begin{tabular}{lcc}\hline
\textbf{Method} &\textbf{Top-1 F1} &\textbf{Top-3 F1} \\\hline
\textbf{\sysname{}} &\textbf{45.74} &\textbf{57.25} \\
\hspace{3mm} w/o paragraph ranking &36.33 &44.84 \\
\hspace{3mm} w/o CRF in the sequence tagger &41.26 &49.57 \\
\bottomrule
\end{tabular}
}
\caption{\footnotesize Results of \sysname{} without auxiliary components for the QuoteR test set.}\label{tab:ablation_analysis}
\end{table}
\noindent\textit{Vector space analysis}: \textcolor{black}{Apparently the two tasks presented inside the multi-task framework may seem similar. To understand how the two tasks are different we conduct an experiment by passing 400 test examples (i.e., context-paragraph-quote triples) from QuoteR data as input to the trained multi-task framework and extracting the last hidden representations of the two task specific heads.  We measure the average cosine similarity between these two representations and observe a very small similarity of 0.23. We also plot the cosine similarity heatmap and T-SNE plot between these two representations (see Figure~\ref{fig:vector_space_analysis}). From the T-SNE plot we observe a negligible overlap, which indicates the tasks are indeed not the same, rather the complementary nature of the two tasks assist each other resulting in improvement of both ``quotability detection and recommendation''.}\\
\noindent\textit{Additional user evaluation}: \textcolor{black}{We already presented a set of human evaluation in section~\ref{sec:analysis_reader} to analyze the top predicted quotes from \textsc{MtR}. Here we randomly select 50 examples each from the Gandhi and Quotus data combined, and collect human judgements from  two undergraduate students about the ‘context relevance’ and ‘quotability’. Given the context and predicted quote, we ask them to select ‘Yes/No’ for the `context relevance' and `quotability'. Overall, the two annotators respectively mark ‘Yes’ in 46, 44 cases for ‘context relevance’, and 47, 41 cases for ‘quotability’. }

\section{Conclusion}
In this work, we proposed a method to recommend quotes from large texts given a context. 
We employed a novel multi-task framework for quote prediction, which can in parallel predict the span of text and identify the quotable phrases. We constructed three datasets of different genres and experimented on them. We believe that our methodology and datasets will be beneficial for future research.

\section*{Acknowledgments}
We sincerely thank Saketh Konda and Agnik Saha for their invaluable contributions to this work. Their assistance in conducting qualitative analyses and running baseline experiments played an integral role in shaping this research. We also extend our gratitude to the anonymous reviewers for their thoughtful feedback and constructive suggestions, which significantly enhanced the quality of this paper.

\section*{Limitations}
In this section we will discuss the limitations of our study. While it is evident that the quotes are available in different regional languages, all of our experiments are conducted for the English version of the datasets. Few of the pre-processing steps might not be suitable for the languages with different morphosyntactic structures. Further the base models will also need to be changed.

\section*{Ethics statement}
We used three datasets for our experiments. The QuoteR dataset was released publicly by the authors of \cite{quoteR}. Besides,
we extracted all the paragraphs from open corpora, including free public domain e-books. The quotes of Gandhi were collected from the free quote repository and the context were extracted from the publicly available portal. Both the quotes and the contexts for the Quotus data were collected from the open corpora. The annotators voluntarily annotated the predictions for our analysis, and we did not retain any of their private information.

\bibliography{custom}

\begin{thebibliography}{34}
\providecommand{\natexlab}[1]{#1}

\bibitem[{Adak et~al.(2020)Adak, Vyas, Mukherjee, Ambavi, Kadasi, Singh, and Patel}]{adak2020gandhipedia}
Sayantan Adak, Atharva Vyas, Animesh Mukherjee, Heer Ambavi, Pritam Kadasi, Mayank Singh, and Shivam Patel. 2020.
\newblock \href {https://doi.org/10.1145/3383583.3398631} {Gandhipedia: A one-stop ai-enabled portal for browsing gandhian literature, life-events and his social network}.
\newblock In \emph{Proceedings of the ACM/IEEE Joint Conference on Digital Libraries in 2020}, JCDL '20, page 539–540, New York, NY, USA. Association for Computing Machinery.

\bibitem[{Bendersky and Smith(2012)}]{bendersky-smith-2012-dictionary}
Michael Bendersky and David Smith. 2012.
\newblock \href {https://aclanthology.org/W12-2510} {A dictionary of wisdom and wit: Learning to extract quotable phrases}.
\newblock In \emph{Proceedings of the {NAACL}-{HLT} 2012 Workshop on Computational Linguistics for Literature}, pages 69--77, Montr{\'e}al, Canada. Association for Computational Linguistics.

\bibitem[{Brown et~al.(2020)Brown, Mann, Ryder, Subbiah, Kaplan, Dhariwal, Neelakantan, Shyam, Sastry, Askell et~al.}]{brown2020language}
Tom Brown, Benjamin Mann, Nick Ryder, Melanie Subbiah, Jared~D Kaplan, Prafulla Dhariwal, Arvind Neelakantan, Pranav Shyam, Girish Sastry, Amanda Askell, et~al. 2020.
\newblock Language models are few-shot learners.
\newblock \emph{Advances in neural information processing systems}, 33:1877--1901.

\bibitem[{Chen et~al.(2017)Chen, Fisch, Weston, and Bordes}]{chen2017reading}
Danqi Chen, Adam Fisch, Jason Weston, and Antoine Bordes. 2017.
\newblock \href {https://doi.org/10.18653/v1/P17-1171} {Reading {W}ikipedia to answer open-domain questions}.
\newblock In \emph{Proceedings of the 55th Annual Meeting of the Association for Computational Linguistics (Volume 1: Long Papers)}, pages 1870--1879, Vancouver, Canada. Association for Computational Linguistics.

\bibitem[{Dai and Callan(2019)}]{dai2019deeper}
Zhuyun Dai and Jamie Callan. 2019.
\newblock \href {https://doi.org/10.1145/3331184.3331303} {Deeper text understanding for ir with contextual neural language modeling}.
\newblock In \emph{Proceedings of the 42nd International ACM SIGIR Conference on Research and Development in Information Retrieval}, SIGIR'19, page 985–988, New York, NY, USA. Association for Computing Machinery.

\bibitem[{Devlin et~al.(2019)Devlin, Chang, Lee, and Toutanova}]{devlin-etal-2019-bert}
Jacob Devlin, Ming-Wei Chang, Kenton Lee, and Kristina Toutanova. 2019.
\newblock \href {https://doi.org/10.18653/v1/N19-1423} {{BERT}: Pre-training of deep bidirectional transformers for language understanding}.
\newblock In \emph{Proceedings of the 2019 Conference of the North {A}merican Chapter of the Association for Computational Linguistics: Human Language Technologies, Volume 1 (Long and Short Papers)}, pages 4171--4186, Minneapolis, Minnesota. Association for Computational Linguistics.

\bibitem[{Gokaslan and Cohen(2019)}]{Gokaslan2019OpenWeb}
Aaron Gokaslan and Vanya Cohen. 2019.
\newblock Openwebtext corpus.
\newblock \url{http://Skylion007.github.io/OpenWebTextCorpus}.

\bibitem[{Grinberg(2018)}]{grinberg2018flask}
Miguel Grinberg. 2018.
\newblock \emph{Flask web development: developing web applications with python}.
\newblock " O'Reilly Media, Inc.".

\bibitem[{Izacard et~al.(2021)Izacard, Caron, Hosseini, Riedel, Bojanowski, Joulin, and Grave}]{izacard2021unsupervised}
Gautier Izacard, Mathilde Caron, Lucas Hosseini, Sebastian Riedel, Piotr Bojanowski, Armand Joulin, and Edouard Grave. 2021.
\newblock Unsupervised dense information retrieval with contrastive learning.
\newblock \emph{arXiv preprint arXiv:2112.09118}.

\bibitem[{Izacard and Grave(2021)}]{izacard2021leveraging}
Gautier Izacard and {\'E}douard Grave. 2021.
\newblock Leveraging passage retrieval with generative models for open domain question answering.
\newblock In \emph{Proceedings of the 16th Conference of the European Chapter of the Association for Computational Linguistics: Main Volume}, pages 874--880.

\bibitem[{Lee et~al.(2016)Lee, Ahn, Lee, Ha, and Lee}]{lee}
Hanbit Lee, Yeonchan Ahn, Haejun Lee, Seungdo Ha, and Sang-goo Lee. 2016.
\newblock \href {https://doi.org/10.1145/2911451.2914734} {Quote recommendation in dialogue using deep neural network}.
\newblock In \emph{Proceedings of the 39th International ACM SIGIR Conference on Research and Development in Information Retrieval}, SIGIR '16, page 957–960, New York, NY, USA. Association for Computing Machinery.

\bibitem[{Lee et~al.(2018)Lee, Yun, Kim, Ko, and Kang}]{lee-etal-2018-ranking}
Jinhyuk Lee, Seongjun Yun, Hyunjae Kim, Miyoung Ko, and Jaewoo Kang. 2018.
\newblock \href {https://doi.org/10.18653/v1/D18-1053} {Ranking paragraphs for improving answer recall in open-domain question answering}.
\newblock In \emph{Proceedings of the 2018 Conference on Empirical Methods in Natural Language Processing}, pages 565--569, Brussels, Belgium. Association for Computational Linguistics.

\bibitem[{Lewis et~al.(2020)Lewis, Perez, Piktus, Petroni, Karpukhin, Goyal, K\"{u}ttler, Lewis, Yih, Rockt\"{a}schel, Riedel, and Kiela}]{NEURIPS2020_6b493230}
Patrick Lewis, Ethan Perez, Aleksandra Piktus, Fabio Petroni, Vladimir Karpukhin, Naman Goyal, Heinrich K\"{u}ttler, Mike Lewis, Wen-tau Yih, Tim Rockt\"{a}schel, Sebastian Riedel, and Douwe Kiela. 2020.
\newblock \href {https://proceedings.neurips.cc/paper\_files/paper/2020/file/6b493230205f780e1bc26945df7481e5-Paper.pdf} {Retrieval-augmented generation for knowledge-intensive nlp tasks}.
\newblock In \emph{Advances in Neural Information Processing Systems}, volume~33, pages 9459--9474. Curran Associates, Inc.

\bibitem[{Liu et~al.(2019)Liu, Pang, and Liu}]{liu-etal-2019-neural-based}
Yuanchao Liu, Bo~Pang, and Bingquan Liu. 2019.
\newblock \href {https://doi.org/10.18653/v1/P19-1552} {Neural-based {C}hinese idiom recommendation for enhancing elegance in essay writing}.
\newblock In \emph{Proceedings of the 57th Annual Meeting of the Association for Computational Linguistics}, pages 5522--5526, Florence, Italy. Association for Computational Linguistics.

\bibitem[{MacLaughlin et~al.(2021)MacLaughlin, Chen, Ayan, and Roth}]{maclaughlin2021context}
Ansel MacLaughlin, Tao Chen, Burcu~Karagol Ayan, and Dan Roth. 2021.
\newblock \href {https://doi.org/10.1609/icwsm.v15i1.18070} {Context-based quotation recommendation}.
\newblock \emph{Proceedings of the International AAAI Conference on Web and Social Media}, 15(1):397--408.

\bibitem[{MacLaughlin and Smith(2021)}]{maclaughlin-smith-2021-content}
Ansel MacLaughlin and David Smith. 2021.
\newblock \href {https://doi.org/10.18653/v1/2021.eacl-main.195} {Content-based models of quotation}.
\newblock In \emph{Proceedings of the 16th Conference of the European Chapter of the Association for Computational Linguistics: Main Volume}, pages 2296--2314, Online. Association for Computational Linguistics.

\bibitem[{Mann and Whitney(1947)}]{mann1947test}
Henry~B Mann and Donald~R Whitney. 1947.
\newblock On a test of whether one of two random variables is stochastically larger than the other.
\newblock \emph{The annals of mathematical statistics}, pages 50--60.

\bibitem[{Niculae et~al.(2015)Niculae, Suen, Zhang, Danescu-Niculescu-Mizil, and Leskovec}]{niculae2015quotus}
Vlad Niculae, Caroline Suen, Justine Zhang, Cristian Danescu-Niculescu-Mizil, and Jure Leskovec. 2015.
\newblock Quotus: The structure of political media coverage as revealed by quoting patterns.
\newblock In \emph{Proceedings of the 24th International Conference on World Wide Web}, pages 798--808.

\bibitem[{Nogueira et~al.(2020)Nogueira, Jiang, Pradeep, and Lin}]{nogueira-etal-2020-document}
Rodrigo Nogueira, Zhiying Jiang, Ronak Pradeep, and Jimmy Lin. 2020.
\newblock \href {https://doi.org/10.18653/v1/2020.findings-emnlp.63} {Document ranking with a pretrained sequence-to-sequence model}.
\newblock In \emph{Findings of the Association for Computational Linguistics: EMNLP 2020}, pages 708--718, Online. Association for Computational Linguistics.

\bibitem[{Nogueira et~al.(2019)Nogueira, Yang, Cho, and Lin}]{nogueira2019multi}
Rodrigo Nogueira, Wei Yang, Kyunghyun Cho, and Jimmy Lin. 2019.
\newblock Multi-stage document ranking with bert.
\newblock \emph{arXiv preprint arXiv:1910.14424}.

\bibitem[{Pareti et~al.(2013)Pareti, O{'}Keefe, Konstas, Curran, and Koprinska}]{pareti-etal-2013-automatically}
Silvia Pareti, Tim O{'}Keefe, Ioannis Konstas, James~R. Curran, and Irena Koprinska. 2013.
\newblock \href {https://aclanthology.org/D13-1101} {Automatically detecting and attributing indirect quotations}.
\newblock In \emph{Proceedings of the 2013 Conference on Empirical Methods in Natural Language Processing}, pages 989--999, Seattle, Washington, USA. Association for Computational Linguistics.

\bibitem[{Portelli et~al.(2021)Portelli, Passab{\`\i}, Lenzi, Serra, Santus, and Chersoni}]{portelli2021improving}
Beatrice Portelli, Daniele Passab{\`\i}, Edoardo Lenzi, Giuseppe Serra, Enrico Santus, and Emmanuele Chersoni. 2021.
\newblock Improving adverse drug event extraction with spanbert on different text typologies.
\newblock In \emph{International Workshop on Health Intelligence}, pages 87--99. Springer.

\bibitem[{Qi et~al.(2022)Qi, Yang, Yi, Cheng, Liu, and Sun}]{quoteR}
Fanchao Qi, Yanhui Yang, Jing Yi, Zhili Cheng, Zhiyuan Liu, and Maosong Sun. 2022.
\newblock Quoter: A benchmark of quote recommendation for writing.
\newblock In \emph{Proceedings of the 60th Annual Meeting of the Association for Computational Linguistics (Volume 1: Long Papers)}, pages 336--348.

\bibitem[{Raffel et~al.(2020)Raffel, Shazeer, Roberts, Lee, Narang, Matena, Zhou, Li, and Liu}]{10.5555/3455716.3455856}
Colin Raffel, Noam Shazeer, Adam Roberts, Katherine Lee, Sharan Narang, Michael Matena, Yanqi Zhou, Wei Li, and Peter~J. Liu. 2020.
\newblock Exploring the limits of transfer learning with a unified text-to-text transformer.
\newblock \emph{J. Mach. Learn. Res.}, 21(1).

\bibitem[{Scheible et~al.(2016)Scheible, Klinger, and Pad{\'o}}]{scheible-etal-2016-model}
Christian Scheible, Roman Klinger, and Sebastian Pad{\'o}. 2016.
\newblock \href {https://doi.org/10.18653/v1/P16-1164} {Model architectures for quotation detection}.
\newblock In \emph{Proceedings of the 54th Annual Meeting of the Association for Computational Linguistics (Volume 1: Long Papers)}, pages 1736--1745, Berlin, Germany. Association for Computational Linguistics.

\bibitem[{Tan et~al.(2018)Tan, Wan, Liu, and Xiao}]{tan}
Jiwei Tan, Xiaojun Wan, Hui Liu, and Jianguo Xiao. 2018.
\newblock Quoterec: Toward quote recommendation for writing.
\newblock \emph{ACM Transactions on Information Systems (TOIS)}, 36(3):1--36.

\bibitem[{Tan et~al.(2015)Tan, Wan, and Xiao}]{tan2015learning}
Jiwei Tan, Xiaojun Wan, and Jianguo Xiao. 2015.
\newblock \href {https://doi.org/10.1609/aaai.v29i1.9530} {Learning to recommend quotes for writing}.
\newblock \emph{Proceedings of the AAAI Conference on Artificial Intelligence}, 29(1).

\bibitem[{Tan et~al.(2016)Tan, Wan, and Xiao}]{10.1145/2983323.2983788}
Jiwei Tan, Xiaojun Wan, and Jianguo Xiao. 2016.
\newblock \href {https://doi.org/10.1145/2983323.2983788} {A neural network approach to quote recommendation in writings}.
\newblock In \emph{Proceedings of the 25th ACM International on Conference on Information and Knowledge Management}, CIKM '16, page 65–74, New York, NY, USA. Association for Computing Machinery.

\bibitem[{Voskarides et~al.(2021)Voskarides, Meij, Sauer, and de~Rijke}]{vosk}
Nikos Voskarides, Edgar Meij, Sabrina Sauer, and Maarten de~Rijke. 2021.
\newblock News article retrieval in context for event-centric narrative creation.
\newblock In \emph{Proceedings of the 2021 ACM SIGIR International Conference on Theory of Information Retrieval}, pages 103--112.

\bibitem[{Wang et~al.(2021)Wang, Zeng, and Wong}]{wang_trans}
Lingzhi Wang, Xingshan Zeng, and Kam-Fai Wong. 2021.
\newblock Quotation recommendation and interpretation based on transformation from queries to quotations.
\newblock In \emph{Proceedings of the 59th Annual Meeting of the Association for Computational Linguistics and the 11th International Joint Conference on Natural Language Processing (Volume 2: Short Papers)}, pages 754--758.

\bibitem[{Wang et~al.(2019)Wang, Ng, Ma, Nallapati, and Xiang}]{wang2019multi}
Zhiguo Wang, Patrick Ng, Xiaofei Ma, Ramesh Nallapati, and Bing Xiang. 2019.
\newblock Multi-passage bert: A globally normalized bert model for open-domain question answering.
\newblock In \emph{Proceedings of the 2019 Conference on Empirical Methods in Natural Language Processing and the 9th International Joint Conference on Natural Language Processing (EMNLP-IJCNLP)}, pages 5878--5882.

\bibitem[{Wu et~al.(2016)Wu, Schuster, Chen, Le, Norouzi, Macherey, Krikun, Cao, Gao, Macherey et~al.}]{wu2016google}
Yonghui Wu, Mike Schuster, Zhifeng Chen, Quoc~V Le, Mohammad Norouzi, Wolfgang Macherey, Maxim Krikun, Yuan Cao, Qin Gao, Klaus Macherey, et~al. 2016.
\newblock Google's neural machine translation system: Bridging the gap between human and machine translation.
\newblock \emph{arXiv preprint arXiv:1609.08144}.

\bibitem[{Yilmaz et~al.(2019)Yilmaz, Wang, Yang, Zhang, and Lin}]{yilmaz2019applying}
Zeynep~Akkalyoncu Yilmaz, Shengjin Wang, Wei Yang, Haotian Zhang, and Jimmy Lin. 2019.
\newblock Applying bert to document retrieval with birch.
\newblock In \emph{Proceedings of the 2019 Conference on Empirical Methods in Natural Language Processing and the 9th International Joint Conference on Natural Language Processing (EMNLP-IJCNLP): System Demonstrations}, pages 19--24.

\bibitem[{Zhu et~al.(2015)Zhu, Kiros, Zemel, Salakhutdinov, Urtasun, Torralba, and Fidler}]{zhu2015aligning}
Yukun Zhu, Ryan Kiros, Rich Zemel, Ruslan Salakhutdinov, Raquel Urtasun, Antonio Torralba, and Sanja Fidler. 2015.
\newblock Aligning books and movies: Towards story-like visual explanations by watching movies and reading books.
\newblock In \emph{Proceedings of the IEEE international conference on computer vision}, pages 19--27.

\end{thebibliography}

\clearpage

\appendix

\section{Ablation for the Multi-task reader}
\label{appendix:ablation_multitask}
\textcolor{black}{In this section we ablate the sequence tagger module of multi-task reader to observe the performance variations of the reader. We use a simple SpanBERT module in the multi-task framework to train on the QuoteR training set. Then we We use the QuoteR test set for running the inference. The result is shown in the Table~\ref{tab:ablation_multi_task_reader}. We observe that the SpanBERT (span prediction) along with SpanBERT-CRF (for sequence tagging) is working the best for the multi-task framework.}

\begin{table}[!htp]\centering
\scriptsize
\begin{tabular}{lcc}\toprule
\textbf{Approach} &\textbf{EM} &\textbf{BoW F1} \\\midrule
\textbf{SpanBERT (span prediction)} & &\\
\hspace{6mm} $\oplus$ SpanBERT-CRF (sequence tagging) &\textbf{77$\pm$0.4} &\textbf{86$\pm$0.2} \\
\hspace{6mm} $\oplus$ SpanBERT (sequence tagging) &74.1$\pm$0.3 &82.3$\pm$0.4 \\
\bottomrule
\end{tabular}
\caption{\footnotesize Ablation for the Multi-task reader. $\oplus$ denotes multi-task framework.}\label{tab:ablation_multi_task_reader}
\end{table}

\section{Results using different LLMs as reader}
\label{appendix:llm_results}
Extending the Table~\ref{tab:end-to-end-inference}, we demonstrates the results while using different other LLMs.

\begin{table*}[h]\centering
\scriptsize
\begin{tabular}{l|cc|cc|ccc}\toprule
\multirow{3}{*}{\textbf{Method}} &\multicolumn{6}{c}{\textbf{Dataset}} \\
&\multicolumn{2}{c|}{\textbf{QuoteR (test)}} &\multicolumn{2}{c|}{\textbf{Gandhi}} &\multicolumn{2}{c}{\textbf{Quotus}} \\
&\textbf{Top-1 F1} &\textbf{Top-3 F1} &\textbf{Top-1 F1} &\textbf{Top-3 F1} &\textbf{Top-1 F1} &\textbf{Top-3 F1} \\\midrule
Vector-store based retriever + LLM reader (FLAN-T5-large) &{14.5} &{19.23} &{18.2} &{22.5} &{25.27} &{31.2} \\
Vector-store based retriever + LLM reader (FLAN-T5-xl) &{13.32} &{19.4} &{16.54} &{24.33} &{35.0} &{38.25} \\
Vector-store based retriever + LLM reader (bloomz-3b) &{10.33} &{12.12} &{9.19} &{13.48} &{16.43} &{21.31} \\
Vector-store based retriever + LLM reader (Falcon-7b) &{10.08} &{13.34} &{17.05} &{22.35} &{29.73} &{36.73} \\
Vector-store based retriever + LLM reader (Llama-3-8b-instruct) &{14.81} &{18.76} &{17.53} &{23.28} &{31.38} &{39.77} \\
\bottomrule
\end{tabular}
\caption{\footnotesize The result for the quote extraction using the different LLMs as reader.}\label{tab:end-to-end-inference_LLM}
\end{table*}

\section{Deployment status}
\begin{figure}[ht]
    \centering
    \fbox{\includegraphics[width=0.95\columnwidth]{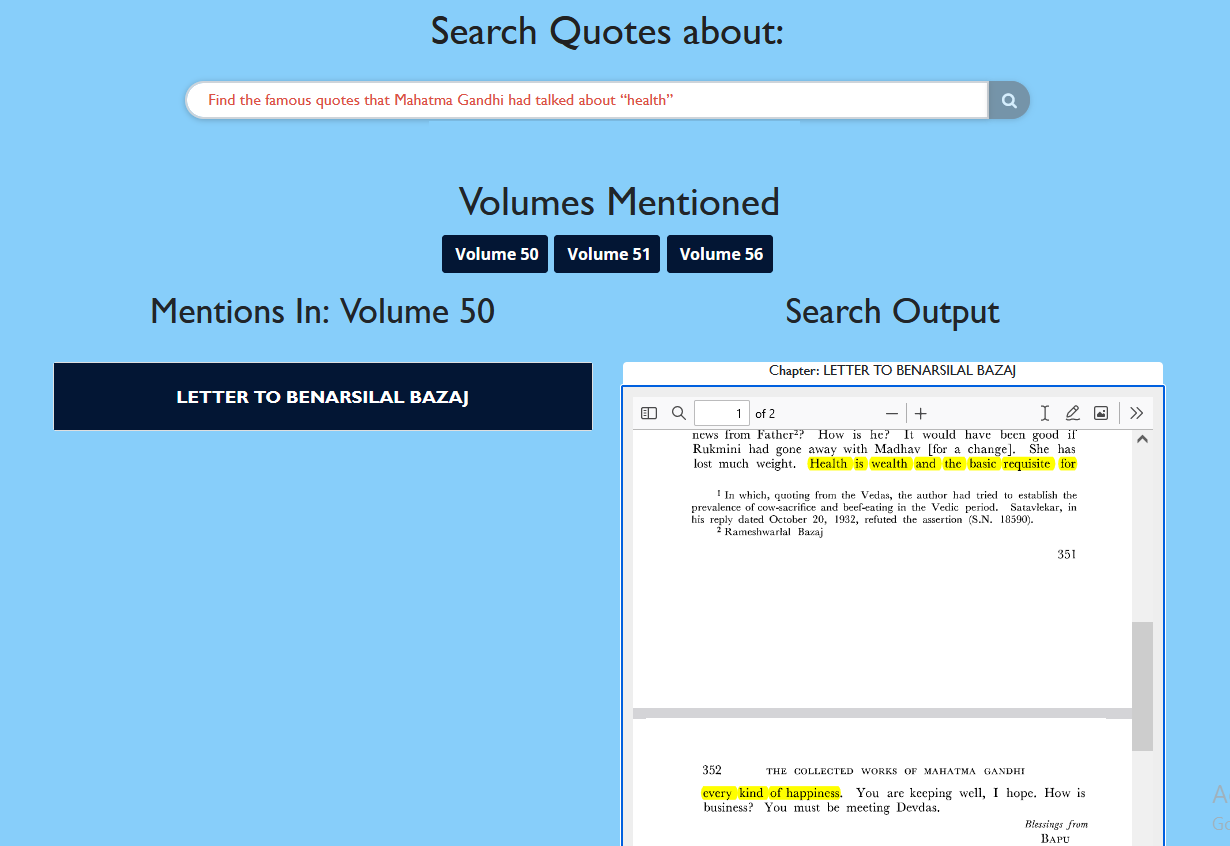}}
    \caption{\footnotesize Example of a real time quote extraction from the Collected Works of Mahatma Gandhi. The output quote is highlighted in \fcolorbox{yellow}{yellow}{yellow} in the pdf.}
    \label{fig:deployment_fig}
\end{figure}

We have deployed the \sysname{} framework in a flask \cite{grinberg2018flask} based web application (link will be made public upon acceptance). We plan to integrate this system with the publicly available and fully searchable historical encyclopedia (e.g., Gandhipedia\footnote{\url{https://gandhipedia150.in/en/}}). We present an example page of our demo system in Figure~\ref{fig:deployment_fig}. The figure shows the result when a user searches for the query ``Find the famous quotes that Mahatma Gandhi had made about \textit{health}''. The system extracts the most relevant quotes from the entire 100 volumes of the Collected Works of Mahatma Gandhi and highlights them in \fcolorbox{yellow}{yellow}{yellow}. 

\begin{figure*}[!ht]
    \centering
    \begin{subfigure}[b]{0.25\textwidth}
        \label{Fig:paragraphs/books}
        \centering
        \includegraphics[width=\textwidth]{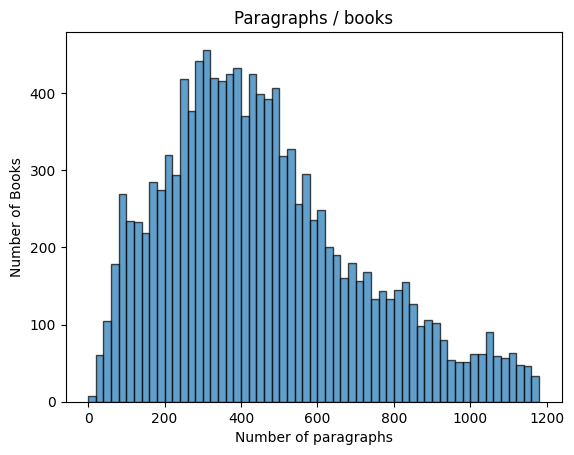}
        \caption{\footnotesize QuoteR}
    \end{subfigure}
    \begin{subfigure}[b]{0.25\textwidth}
        \label{Fig:paragraphs/books_gandhi}
        \centering
        \includegraphics[width=\textwidth]{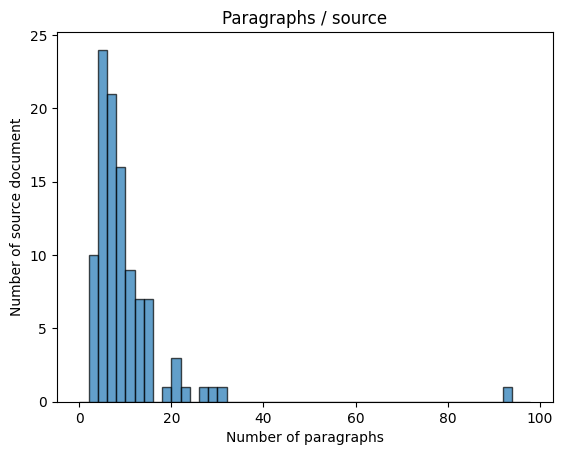}
        \caption{\footnotesize Gandhi}
    \end{subfigure}
    \begin{subfigure}[b]{0.25\textwidth}
        \label{Fig:paragraphs/books_quotus}
        \centering
        \includegraphics[width=\textwidth]{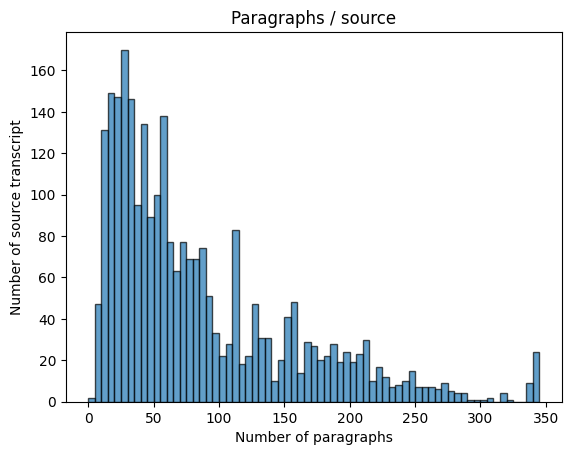}
        \caption{\footnotesize Quotus}
    \end{subfigure}
    \caption{\footnotesize \# of paragraphs per source documents.}
    \label{fig:data_statistics1}
\end{figure*}
\begin{figure*}[!ht]
    \centering
    \begin{subfigure}[b]{0.25\textwidth}
        \label{Fig:tokens/quotes}
        \centering
        \includegraphics[width=\textwidth]{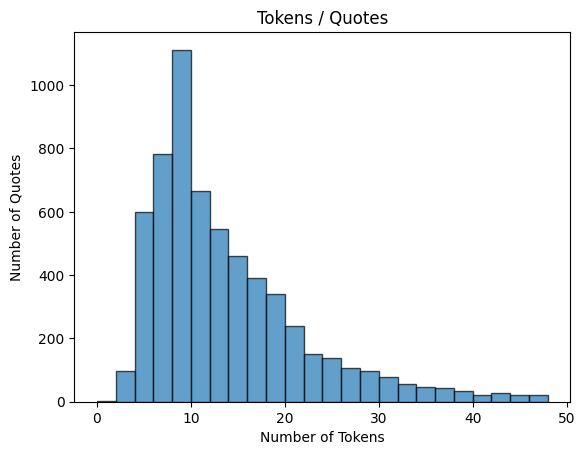}
        \caption{\footnotesize QuoteR}
    \end{subfigure}
    \begin{subfigure}[b]{0.25\textwidth}
        \label{Fig:tokens/quotes_gandhi}
        \centering
        \includegraphics[width=\textwidth]{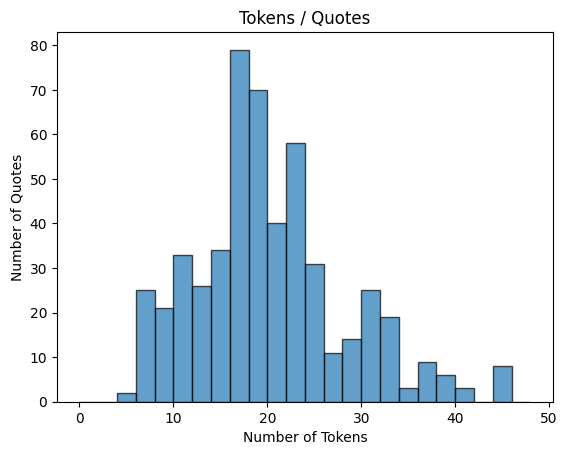}
        \caption{\footnotesize Gandhi}
    \end{subfigure}
    \begin{subfigure}[b]{0.25\textwidth}
        \label{Fig:tokens/quotes_quotus}
        \centering
        \includegraphics[width=\textwidth]{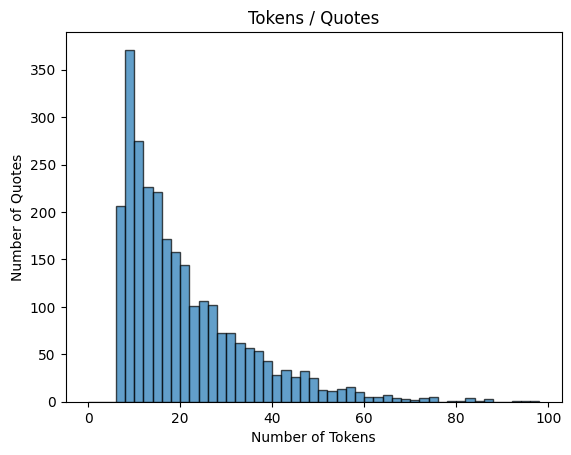}
        \caption{\footnotesize Quotus}
    \end{subfigure}
    \caption{\footnotesize Distributions over source document, paragraphs, and quote lengths.}
     \label{fig:data_statistics2}
     \vspace{1em}
\end{figure*}

\section{Data preprocessing details}
\label{appendix:data_preprocessing}
\paragraph{QuoteR data}: The \textit{Project Gutenberg corpus} comprises more than 73000 e-books in textual form. We assign each of these books a unique bookID and divide each book into fixed-length (i.e., 200 word length) paragraphs, and assign each such paragraph a unique paragraphID. The distribution of the number of paragraphs per book and the number of tokens per quote is presented in Figure~\ref{fig:data_statistics1} and \ref{fig:data_statistics2}. We construct a TF-IDF weighted word-doc sparse matrix \cite{chen2017reading} from all the documents, index, and store this content in the \texttt{sqlite db}. For each of the quotes present in the QuoteR dataset we recursively search for the appearance of the quote in each of these books. Once a search gets a hit, we link the bookID with that particular quote (to be used for training the paragraph retrieval model). Since the authors in \cite{quoteR} stored the context from different sources and the correct mapping to the books is not present, we consider the 40 words
preceding and following it as its left and right
contexts, respectively. Similar to \cite{quoteR} the concatenation of the left and right contexts forms a complete context. We then store the context, quote, and positive paragraph (to be used for training the quote prediction model). Out of the 6108 unique quotes, we are able to find the occurrences of 1708 quotes from the \textit{Project Gutenberg} and we finally construct 4889 quote-context-paragraph (one quote may contain multiple contexts) triples as examples for training and evaluating. The algorithm for generating the quote-context-paragraph triples is presented in Algorithm~\ref{algo:data_preparation}.\\
\paragraph{Gandhi data}: Similarly, for the Gandhi quotes, we search for the quotes in the CWMG and find their appearance in a particular chapter of a book in the CWMG. We utilize the \textit{Gandhipedia}~\cite{adak2020gandhipedia} engine, which uses an elasticsearch based search engine to locate the quotes. We consider the 40 preceding and following words from each quote in the particular chapter as its context. In addition, we find that out of all the Gandhi quotes, three quotes are already present in the QuoteR set. We, therefore, remove them from the Gandhi data. Finally, we obtain 737 quote-context-paragraph triples.

\paragraph{Generating target context}: Unlike the Quotus data, we do not have explicit target documents (i.e., where the quote needs to be recommended from source) for the QuoteR and Gandhi dataset. We synthesize the target context by paraphrasing the original context in the corresponding dataset. This is performed to reduce the overlapping words of the target context and the source document and to effectively evaluate the robustness of the methodology. We use ChatGPT\footnote{\url{https://openai.com/blog/chatgpt}} API to paraphrase the context. The examples of such paraphrased context are provided in Appendix~\ref{examples_paraphrased}. The prompt used for paraphrasing:\\
\texttt{As a paraphrasing expert can you rephrase the following input text? Ensure the rephrased text incorporates a differ-\\ent range of vocabulary compared to the original text.\\
Input text: \{<Input text>\}\\
Rephrased text: }\\
To analyse the hardness of the generated target contexts, we measure the word overlap between the original context and the rephrased context. We observe that the average word overlap ratio between the original and the rephrased contexts are - 0.19 and 0.18 for QuoteR and Gandhi data respectively. This indicates that the rephrased target context has significantly different words thus making the task of the paragraph retriever harder. We also measure whether the meaning of the rephrased contexts get significantly deviated from the original context. We use \textit{GPT-4} to calculate the widely used \textit{Faithfulness} of the rephrased contexts with respect to the original context. We observe an average \textit{Faithfulness} of 0.84 and 0.88 for the QuoteR and Gandhi data respectively, which ensures that the rephrased contexts preserves the meaning of original contexts.

\paragraph{Quotus data}: For the Quotus dataset, we utilize the Quoting POTUS website\footnote{\url{http://snap.stanford.edu/quotus/vis/}} to collect a set of examples for our experiments. They release the transcripts and the collection of aligned quotes, containing
the text of the quote in the news article, its aligned position within the source transcript, and the corresponding news article metadata (title, url, timestamp). We crawl the provided news article URLs and extract the body content of each news article using BeautifulSoup\footnote{https://www.crummy.com/software/BeautifulSoup/bs4/doc/}. We are able to extract 10,114 news articles in this way (some of the links were not working and could not be crawled). To locate the quotes within the news articles, we utilize regular expressions and identify the appearance of 2,698 quotes. We then consider the 40 preceding and following words from each quote in the news article as its context. In the released dataset, the source transcript is already divided into several paragraphs, and the alignment of the quotes to the positive paragraph is also provided. As a result, we did not need to explicitly create the quote-paragraph alignment. This yields a total of 2,698 quote-context-paragraph triples, which we use for our experiments.

The Algorithm~\ref{algo:data_preparation} shows the step-by-step procedure to prepare the dataset for our experiments. The auxiliary functions (i.e., Algorithms~\ref{algo:segment_book}, \ref{algo:generate_context} and \ref{algo:quote_book_mapping}) used in the algorithms are depicted in the subsequent algorithms. 

\begin{algorithm*}
\caption{\footnotesize Paragraph retrieval data generation}
\label{algo:data_preparation}
\begin{algorithmic}[1]
    \Require{$\text{list\_of\_quotes}$: list of selected quotes; $\text{corpus\_directory}$: directory of the corpus (ex. Gutenberg)}
    \State $quoteid\_book\_mapping \gets$ \newline $CREATE\_QUOTE\_TO\_BOOK\_MAPPING(list\_of\_quotes, corpus\_directory)$
    \State $ctxid \gets 0$ \Comment{Initialize Context Id}
    \State $ctxid\_to\_text \gets \{\}$ \Comment{Initialize Context Id to Context text mapping}
    \State $quoteid\_to\_ctxid \gets \{\}$ \Comment{Initialize Quote Id to Context Id mapping}
    \ForAll{$(quoteid, \text{list\_of\_book\_paths})$ \textbf{in} $quoteid\_book\_mapping$}
        \State $dataset \gets []$ \Comment{Dataset to be used for training and testing paragraph retrieval}
        \State $quoteid\_to\_ctxid[quoteid] \gets []$
        \ForAll{$book\_path$ \textbf{in} $list\_of\_book\_paths$}
            \State $paragraphs \gets \text{SEGMENT\_BOOK}(book, paragraph_lenght=200)$ \Comment{Segment the book contents into several paragraphs}
            \State $\text{save}(docid\_to\_text)$
            \ForAll{$paragraph$}
                \If{$quote$ \textbf{in} $paragraph$}
                    \State $ctx \gets \text{CREATE\_CONTEXT}(quote, paragraph)$ \Comment{Creating context for a quote}
                    \State $ctxid\_to\_text[ctxid] \gets ctx$
                    \State $dataset.\text{append}([ctxid, [\text{pos\_para\_id}], [\text{candidate\_id}]])$
                    \State $ctxid \gets ctxid + 1$
                    \If{$quoteid$ \textbf{in} $quoteid\_to\_ctxid.\text{keys}()$}
                        \State $quoteid\_to\_ctxid[quoteid].\text{append}(ctxid)$
                    \Else
                        \State $quoteid\_to\_ctxid[quoteid] \gets [ctxid]$
                    \EndIf
                \EndIf
            \EndFor
            \State $\text{save}(dataset)$
        \EndFor
    \EndFor
    \State $\text{save}(ctxid\_to\_text)$
    \State $\text{save}(quoteid\_to\_ctxid)$
\end{algorithmic}
\label{algo1}
\end{algorithm*}

\begin{algorithm*}
\caption{\footnotesize Create quote to book mapping}
\label{algo:quote_book_mapping}
\begin{algorithmic}[1]
    \Function{CREATE\_QUOTE\_TO\_BOOK\_MAPPING}{$\text{list\_of\_quotes, corpus\_directory}$}
        \State \textbf{Input:} $\text{list\_of\_quotes, corpus\_directory}$
        \State \textbf{Output:} $\text{quote\_to\_book\_mapping}$
        
        \State $\text{quote\_to\_book\_mapping} \gets \{\}$
        
        \ForAll{$\text{quote}$ \textbf{in} $\text{list\_of\_quotes}$}
            \ForAll{$\text{book\_path}$ \textbf{in} $\text{corpus\_directory}$}
                \If{$\text{quote}$ \text{found in} $\text{book\_path}$}
                    \If{$\text{quote}$ \textbf{in} $\text{quote\_to\_book\_mapping}$}
                        \State $\text{quote\_to\_book\_mapping[quote].append(book\_path)}$
                    \Else
                        \State $\text{quote\_to\_book\_mapping[quote]} \gets [\text{book\_path}]$
                    \EndIf
                \EndIf
            \EndFor
        \EndFor
        
        \State \textbf{return} $\text{quote\_to\_book\_mapping}$
    \EndFunction
\end{algorithmic}
\end{algorithm*}

\begin{algorithm*}
\caption{\footnotesize Segment book into paragraphs of fixed length}
\label{algo:segment_book}
\begin{algorithmic}[1]
    \Function{segment\_book}{$\text{text\_document}, \text{paragraph\_length}$}
        \State \textbf{Input:} $\text{text\_document}, \text{paragraph\_length}$
        \State \textbf{Output:} $paragraphs$
        
        \State $paragraphs \gets \{\}$
        \State $current\_paragraph \gets ""$
        \State $current\_paragraph\_id \gets 0$
        
        \For{$word$ \textbf{in} $text\_document.split()$}
            \State $current\_paragraph \gets current\_paragraph + " " + word$
            
            \If{$\text{len}(current\_paragraph) \geq \text{paragraph\_length}$}
                \State $paragraphs[current\_paragraph\_id] \gets \text{current\_paragraph.strip()}$
                \State $current\_paragraph \gets ""$
                \State $current\_paragraph\_id \gets current\_paragraph\_id + 1$
            \EndIf
        \EndFor
        
        \If{$\text{len}(current\_paragraph) > 0$}
            \State $paragraphs[current\_paragraph\_id] \gets \text{current\_paragraph.strip()}$
        \EndIf
        
        \State \textbf{return} $paragraphs$
    \EndFunction
\end{algorithmic}
\end{algorithm*}

\begin{algorithm*}
\caption{\footnotesize Generate context for a quote in a paragraph}

\label{algo:generate_context}
\begin{algorithmic}[1]
    \Function{CREATE\_CONTEXT}{$\text{quote, paragraph}$}
        \State \textbf{Input:} $\text{quote, paragraph}$
        \State \textbf{Output:} $\text{context}$
        
        \State $\text{context} \gets ""$
        \State $\text{quote\_position} \gets \text{paragraph.find(quote)}$
        
        \If{$\text{quote\_position} \neq -1$}
            \State $\text{preceding\_40} \gets \text{paragraph[:quote\_position].split(" ")[-40:]}$
            \State $\text{following\_40} \gets \text{paragraph[quote\_position + \text{len}(quote):].split(" ")[:40]}$
            \State $\text{context} \gets \text{" ".join(preceding\_40)} \text{ } \text{" ".join(following\_40)}$
        \EndIf
        
        \State \textbf{return} $\text{context}$
    \EndFunction
\end{algorithmic}

\end{algorithm*}

\section{Baseline methods}
\label{appendix:baseline}
\noindent\textbf{Baselines}: Both the retriever and the reader can have many variants which serve as ideal baselines. In the retriever part we use vanilla BM25 as a first baseline. Apart from the simple BM25 retriever, we employ BERT and T5 based re-ranking to improve paragraph selection. For input to BERT we tokenize the contexts and
source document paragraphs into wordpieces~\cite{wu2016google} and cap them at predetermined lengths chosen as hyperparameters. BERT uses a special token [SEP] to separate paragraph from the context. So the final wordpiece input to the BERT is:
$$ [CLS] \quad context \quad [SEP] \quad paragraph \quad [SEP] $$
Following~\cite{wang2019multi}, we fine-tune BERT-base using the pairwise loss. Thus, a single training example for paragraph BERT consists of $n + 1$ instances, i.e., one positive instance plus $n$ negative instances. Each of the $n + 1$ packed input sequences are fed to BERT independently. We use the final hidden vector $\mathbf{C} \in \mathbb{R}^h$ corresponding to the first input token [CLS] as the representation for each of the $n + 1$ sequences, where $h$ is the size of the final hidden layer. In addition, we also fine-tune encoder-decoder based (T5, FLAN-T5) and decoder-only (Llama-3) re-ranking models in the same way as discussed in section~\ref{method:retriever}.\\
For the reader part, as primitive baselines, we consider using the first and last sentences of each paragraph as potential quotes. To further explore, we also fine-tune the BERT and the SpanBERT pretrained models on the BERT question answering architecture. We keep the same hyperparameter settings as the multi-task framework. Again, we fine-tune on 80\% of the QuoteR data, and use 10\% for validation before testing on the remaining 10\%. In addition, we conduct similar few-shot experiments with the Gandhi and the Quotus dataset.\\
\noindent\textit{LLM based baselines}: With the advancement of large language models (LLMs) such as T5 \cite{10.5555/3455716.3455856}, GPT-3 \cite{brown2020language} it is important to observe their ability to perform the task of quote extraction. These models have proven to be highly valuable for contextual learning when provided with specific prompts in zero-shot scenarios. We replace the multi-task reader with different medium sized open-source LLMs such as FLAN-T5-large\footnote{\url{https://huggingface.co/google/flan-t5-large}}, FLAN-T5-XL\footnote{\url{https://huggingface.co/google/flan-t5-xl}}, Bloomz-3b\footnote{\url{https://huggingface.co/bigscience/bloomz-3b}}, Falcon-7b\footnote{\url{https://huggingface.co/tiiuae/falcon-7b}}, Llama-3-8b\footnote{\url{https://huggingface.co/ meta-llama/Meta-Llama-3-8B-Instruct}} models to predict the most relevant quote given the paragraph and context. We use the below prompt:
\begin{mdframed}
    You are an AI assistant in recommending a suitable 'quote' based on the context and your task is to extract a relevant quote from the given pargraph based on the context. Note that, the context and the paragraph may contain grammatical errors. DO NOT use any external information.\\

\noindent Context: "\{context\}"\\

\noindent Paragraph: "\{paragraph\}"\\

\noindent Just extract the relevant quote without any other sentence: 
\end{mdframed}

\section{Model implementation details}
\label{appendix:model_impl}
\paragraph{Retriever}: For retriever we use \textit{lagchain API}\footnote{\url{https://python.langchain.com/docs/modules/data\_connection/}}, employ recursive\_text\_splitter\footnote{\url{https://python.langchain.com/docs/modules/data\_connection/document\_transformers/recursive\_text\_splitter}} for splitting the document, \textit{chromaDB} as vector store. 
For fine-tuning the reranking models we use huggingface API \footnote{\url{https://huggingface.co/}}.\\
\noindent\textit{FLAN-T5}: We fine-tune our T5 models (base\footnote{\url{https://huggingface.co/t5-base}}, large\footnote{\url{https://huggingface.co/t5-large}}) and \textit{FLAN-T5-large}\footnote{\url{https://huggingface.co/google/flan-t5-large}} with a learning rate of 2\(e^{-5}\) and a weight decay of 0.01 for a maximum of 10 epochs with a batch size of 4.
We use a maximum of 1024 input tokens and one output token. Training T5 base, large, and Flan-T5-large take approximately 2, 5, and 6 hours overall, respectively, on a single RTX 4090 GPU.
We use greedy decoding during inference and used \textit{output\_logits=True} while generating text to retrieve unprocessed probabilities assigned to a token. We use  same hyperparameter setting for \textit{Llama-3-8b-instruct}\\
\noindent \textit{bert-base}: For fine-tuning  \textit{bert-base}\footnote{\url{https://huggingface.co/bert-base-uncased}} for the paragraph retrieval task,  we search over a  batch-size \(\in \{4, 8, 16\} \), and set the learning rate of 2\(e^{-5}\). We set the maximum number of epochs to 10. We also perform a search over $n$ \(\in \{3, 6, 9, 12\} \) sampled negative paragraphs per positive paragraph for our paragraph ranking model. We select the best model using the dev set and the best paragraph model is trained with 9 negative examples and a batch size of 16. We used single NVIDIA Tesla P100 GPU for training the model.

\paragraph{Reader}: For the span selection model (the multi-task and other transformers based baseline models), we cap the total length of the context and paragraph to 384 length wordpieces. In case the total length exceeds the maximum length (i.e., 384), we only truncate the paragraph. Similarly, for the quotable phrase identification model (i.e., the sequence tagger model in the multi-task setting) we select a maximum length of 384. We fine-tune the publicly available \textit{spanbert-large}\footnote{\url{https://huggingface.co/SpanBERT/spanbert-large-cased}}, by setting the batch-size \(\in \{4, 8\} \), learning rate of 2\(e^{-5}\). We fine-tune the model over 10 epochs and use early stopping based on the dev set. Again we used single NVIDIA Tesla P100 GPU for training the model. For the multi-task framework, the training process took 3.5 hours to complete.
For the LLM inference we use single NVIDIA Tesla P100 GPU. Additionally, we applied 4bit quantization while loading the larger LLMs as those models would not fit in our GPU.

\section{Examples of paraphrased context}
\label{examples_paraphrased}
Table~\ref{tab:paraphrased_context} shows one paraphrased example from QuoteR and Gandi dataset which were used as the target context. Quotus dataset having a separate target article, we did not require paraphrasing the context. 

\begin{table*}[!ht]\centering

\scriptsize
\begin{tabular}{|l|p{0.4\linewidth}|p{0.4\linewidth}|}\hline
\textbf{Dataset} &\textbf{Actual Context} &\textbf{Paraphrased Context} \\\hline
\textbf{QuoteR} &and for the great Peasant Revolt of 1381. John Ball's famous rhyme condensed the scorn for the nobles, the longing for just rule, and the resentment at oppression, of the peasants of that time and of all times:-- " A hundred years after the Black Death the wages of a common English laborer--we have the highest authority for the statement--commanded twice the amount of the necessaries of life which could have been obtained for the wages paid under &For the significant Peasant Revolt of 1381, John Ball's renowned rhyme encapsulated disdain for the nobles, the yearning for fair governance, and resentment towards oppression. A century after the Black Death, the wages of an ordinary English laborer, as verified by the highest authority, were double the necessities obtainable with previous wages. \\\hline
\textbf{Gandhi} &For, highest perfection is unattainable without highest restraint. Suffering is thus the badge of the human tribe. The goal ever recedes from us. The greater the progress, the greater the recognition of our unworthiness. Full effort is full victory. Therefore, though I realize more than ever how far I am from that goal, for me the Law of complete Love is the law of my being. Each time I fail, my effort shall be &The pinnacle of perfection requires the utmost restraint, and suffering becomes the emblem of the human experience. The goal remains elusive, and progress accentuates our sense of unworthiness. Full effort equates to complete victory. Despite realizing the vast distance from the goal, the Law of complete Love governs my existence. Each failure only strengthens my resolve. \\\hline
\end{tabular}
\caption{\footnotesize Examples of paraphrased contexts for QuoteR and Gandhi datasets.}\label{tab:paraphrased_context}
\end{table*}

\section{Examples of LLM generated quotes}
In Table~\ref{tab:llm_generated_quotes_example} we provide examples of quotes extracted by different LLMs used in our experiments for a specific context and paragraph. We observe that, larger models (such as FLAN-T5-XL, Llama-3-8b)  generate better quotes compared to the smaller models. However, Llama-3 merges some part of the context (``Sweet dewdrops'') in the predicted quote. This is one of the precise reasons why standalone LLMs cannot be reliably used in the quote extraction task, as it may augment the original text which could affect the sanctity of the quote.

\begin{table*}[h]\centering
\scriptsize
\begin{tabular}{c|c|c|l|p{0.2\textwidth}}\toprule
\textbf{Context} &\textbf{Paragraph} &\textbf{Actual quote} &\textbf{Used LLM} &\textbf{Extracted quote} \\\midrule
\multirow{5}{*}{\parbox{0.15\textwidth}{Tonight, the heavens and the earth will mourn your passing, as sweet dewdrops fall from the sky. Your life is fleeting, and your time is coming to an end. The rose, with its}} &\multirow{5}{*}{\parbox{0.3\textwidth}{earth and sky, Sweet dews shall weep thy fall to-night, For thou must die. Sweet rose, whose hue, angry and brave, Bids the rash gazer wipe his eye, Thy root is ever in its grave, And thou must die. Sweet spring, full of sweet days and roses, a box where sweets compacted lie , My music shows you have your closes, And all must die. Only a sweet and virtuous soul, Like seasoned timber, never gives But when the whole world turns to coal, Then chiefly lives. George Herbert. THE LIE. Go, Soul,}} &\multirow{5}{*}{\parbox{0.15\textwidth}{Sweet spring, full of sweet days and roses, a box where sweets compacted lie}} &Bloomz-3b &earth and sky, Sweet dews shall weep thy \\\cline{4-5}
 & & &FLAN-T5-large &rage, is ever in its grave \\\cline{4-5}
 & & &FLAN-T5-XL &Sweet dews shall weep thy fall to-night, For thou must die. Sweet rose, whose hue, angry and brave, Bids the rash \\\cline{4-5}
 & & &Falcon-7b &Sweet dewdrops fall from the sky \\\cline{4-5}
 & & &Llama-3-8b &Sweet dewdrops fall from the sky. For thou must die. Sweet rose, whose hue, angry and brave, Bids the rash gazer wipe his eye, Thy root is ever in \\
\bottomrule
\end{tabular}
\caption{Quotes extracted by different LLMs used for a specific context and paragraph}
\label{tab:llm_generated_quotes_example}
\end{table*}

\end{document}